\pdfoutput=1

\documentclass[11pt]{article}

\usepackage[]{emnlp2021}

\usepackage{graphicx}
\usepackage{listings}
\usepackage{amsmath}
\usepackage{float}
\usepackage{longtable}
\usepackage{tabu}
\usepackage{enumitem}
\usepackage{tabularx}         
\usepackage{booktabs}         
\usepackage{multirow}         
\usepackage{subcaption}
\usepackage{threeparttable}
\usepackage{amssymb}
\usepackage[normalem]{ulem}
\usepackage{xcolor}

\definecolor{codegreen}{rgb}{0,0.6,0}
\definecolor{codegray}{rgb}{0.5,0.5,0.5}
\definecolor{codepurple}{rgb}{0.58,0,0.82}
\definecolor{backcolour}{rgb}{0.95,0.95,0.92}

\lstdefinestyle{mystyle}{
    backgroundcolor=\color{backcolour},   
    commentstyle=\color{codegreen},
    keywordstyle=\color{magenta},
    numberstyle=\tiny\color{codegray},
    stringstyle=\color{codepurple},
    basicstyle=\ttfamily\footnotesize,
    breakatwhitespace=false,         
    breaklines=true,                 
    captionpos=b,                    
    keepspaces=true,                 
    numbers=left,                    
    numbersep=5pt,                  
    showspaces=false,                
    showstringspaces=false,
    showtabs=false,                  
    tabsize=2
}

\usepackage{times}
\usepackage{latexsym}

\usepackage[T1]{fontenc}

\usepackage[utf8]{inputenc}
\DeclareUnicodeCharacter{FFFD}{ }
\usepackage{microtype}

\title{Pushing on Text Readability Assessment: \\A Transformer Meets Handcrafted Linguistic Features}

\author{Bruce W. Lee$^{1,3}$ \\
  Univ. of Pennsylvania$^1$ \\
  PA, USA \\
  \texttt{brucelws@seas.upenn.edu} \\\And
  Yoo Sung Jang$^{2,3}$ \\
  Univ. of Wisconsin-Madison$^2$ \\
  WI, USA \\
  \texttt{yjang43@wisc.edu} \\\And
  Jason Hyung-Jong Lee$^3$ \\
  LXPER AI$^3$ \\
  Seoul, South Korea \\
  \texttt{jasonlee@lxper.com} \\}

\begin{document}{}
\maketitle{}{}
\begin{abstract}
We report two essential improvements in readability assessment: 1. three novel features in advanced semantics and 2. the timely evidence that traditional ML models (e.g. Random Forest, using handcrafted features) can combine with transformers (e.g. RoBERTa) to augment model performance. First, we explore suitable transformers and traditional ML models. Then, we extract 255 handcrafted linguistic features using self-developed extraction software. Finally, we assemble those to create several hybrid models, achieving state-of-the-art (SOTA) accuracy on popular datasets in readability assessment. The use of handcrafted features help model performance on smaller datasets. Nota-bly, our RoBERTA-RF-T1 hybrid achieves the near-perfect classification accuracy of 99\%, a 20.3\% increase from the previous SOTA.

\end{abstract}

\section{Introduction}
The long quest for advancing readability assessment (RA) mostly centered on handcrafting the linguistic features that affect readability \citep{Pitler:08}. RA is a time-honored branch of natural language processing (NLP) that quantifies the difficulty with which a reader understands a text \citep{Feng:10}. Being one of the oldest systematic approaches to linguistics \citep{Collins-Thompson:14}, RA developed various linguistic features. These range from simple measures like the average count of syllables to those as sophisticated as semantic complexity \citep{buchanan2001characterizing}.

Perhaps due to the abundance of dependable linguistic features, an overwhelming majority of RA systems are Support Vector Machines (SVM) with handcrafted features \citep{Hansen:21}. Such traditional machine learning (ML) methods were linguistically explainable, expandable, and most importantly, competent against the modern neural models. As a fragmentary example, \citet{Fili:19} reports that a large ensemble of 6 BiLSTMs with BERT \citep{bert}, ELMo \citep{elmo}, Word2Vec \citep{word2vec}, and GloVe \citep{glove} embeddings showed only $\sim$1\% accuracy improvement from a single SVM model developed by \citet{Xia:16}.

Even though deep neural networks have achieved state-of-the-art (SOTA) performance in almost all semantic tasks where sufficient data were available \citep{colbert:11, Zhang:15}, neural models started showing promising results in RA only quite recently \citep{Martinc:21}. A known challenge for the researchers in RA is the lack of large public datasets -- with the unique exception of WeeBit \citep{Vajjala:12}. Technically speaking, even WeeBit is not entirely public since it has to be directly obtained from the authors.

\citet{Martinc:21} raised the SOTA classification accuracy on the popular WeeBit dataset \citep{Vajjala:12} by about 4\% using BERT. This was the first solid proof that neural models with auto-generated features can show significant improvement compared to traditional ML with handcrafted features. However, neural models, or transformers (which is the interest of this paper), still show not much better performance than traditional ML on smaller datasets like OneStopEnglish \citep{Vajjala:18}, despite the complexity. 

From our observations, the reported low performances of transformers on small RA datasets can be accounted for two reasons. 1. Only BERT was applied to RA, and there could be other transformers that perform better, even on small datasets. 2. If a transformer shows weak performance on small datasets, there must be some additional measures done to supply the final model (e.g. ensemble) with more linguistic information, but such a study is rare in RA. Hence, we tackle the abovementioned issues in this paper. In particular, we 1. perform a wide search on transformers, traditional ML models, and handcrafted features \& 2. develop a hybrid architecture for SOTA and robustness on small datasets.

However, before we move on to hybrid models, we begin by supplementing an underexplored linguistic branch of handcrafted features. According to survey research on RA \citep{Collins-Thompson:14}, the study on advanced semantics is scarce. We lack a model to capture how deeper semantic structures affect readability. We attempt to solve this issue by viewing a text as a collection of latent topics and calculating the probability distribution.

Then, we move on to combine traditional ML (w handcrafted features\footnote{We use ``handcrafted features'' and ``linguistic features'' interchangeably throughout this paper.}) and transformers. Such a hybrid system is only reported by \citet{deutsch-etal-2020-linguistic}, concluding, ``(hybrid models) did not achieve SOTA performance.'' But we obtain contrary results. Through a large study on the optimal combination, we obtain SOTA results on WeeBit and OneStopEnglish. Also, our BERT-GB-T1 hybrid beats the (previous) SOTA accuracy with only 30\% of the full dataset, in section 4.7.

Our main objectives are creating advanced semantic features and hybrid models. But our contributions to academia are not limited to the abovementioned two. We make the following additions:

\noindent
\textbf{1.} We numerically represent certain linguistic properties pertaining to advanced semantics.

\noindent
\textbf{2.} We develop a large-scale, openly available 255 features extraction Python toolkit (which is highly scarce\footnote{An exception is Dr. Vajjala's Java toolkit, available at bitbucket.org/nishkalavallabhi/complexity-features.} in RA). We name the software \textbf{LingFeat\footnote{github.com/brucewlee/lingfeat}}.

\noindent
\textbf{3.} We conduct wide searches and parametrizations on transformers\footnote{github.com/yjang43/pushingonreadability\_transformers} and traditional ML\footnote{github.com/brucewlee/pushingonreadability\_traditional\_ML} for RA use.

\noindent
\textbf{4.} We develop hybrid models for SOTA and robu-stness on small datasets. Notably, RoBERTa-RF-T1 achieves 99\% accuracy on OneStopEnglish, 20.3\% higher than the previous SOTA (table 5).

\section{Advanced Semantics}
\subsection{Overview}
A text is a communication between author and reader, and its readability is affected by the reader having shared world/domain knowledge. According to \citet{Collins-Thompson:14}, the features resulting from topic modeling may characterize the deeper semantic structures of a text. These deeper representations accumulate and appear to us in the form of perceivable properties like sentiment and genre. But advanced semantics aims to capture the deeper representation itself.

Among the four branches of linguistic properties (in RA) identified by \citet{Collins-Thompson:14}, advanced semantics remain unexplored. Lexico-semantic \citep{Lu:11,Malvern:12}, syntactic \citep{heilman-etal-2007-combining,Petersen:09}, and discourse-based \citep{Mcnamara:10} properties had several notable works but little dealt with advanced semantics as the given definition. The existing examples in higher-level semantics focus on word-level complexity \citep{collins2004language, crossley2008assessing, landauer2011word, nam2017predicting}.

Such a phenomenon is complex. The lack of investigation on advanced semantics could be due to its low correlation with readability. This is plausible because RA studies often test their features on a human-labeled dataset, potentially biased towards easily recognizable surface-level features \citep{evans2006cognitive}. Such biases could cause low performance.

Further, it must be noted that: 1. world knowledge might not always directly indicate difficulty, and 2. there can be other existing substitute features that capture similar properties on a word level.

\noindent
S1) \textit{Kindness is good.}

\noindent
S2) \textit{Christmas is good.}

\noindent
S3) \textit{I return with the stipulation to dismiss Smith's} \hspace*{5mm} \textit{case; the same being duly executed by me.}

S2 above seems to require more world knowledge than S1. However, ``Christmas'', as a familiar entity, seems to have no apparent contribution to increased difficulty. If any, similar properties can be captured by word frequency/familiarity measures (lexico-semantics) in a large representative corpus \citep{gondy:13}. Also, it seems that S3 is the most difficult, and this can be easily deduced using entity counts (discourse). Entities mostly introduce conceptual information \citep{Feng:10}.

Our key objective in studying advanced semantics is to identify features that add orthogonal information. In other words, we hope to see a performance increase in our overall RA model rather than specific features' high correlations with readability. 

Given the considerations, we draw two guidelines: 1. develop passage-level features since most word-level attributes are captured by existing features, and 2. value the orthogonal addition of information, not individual feature's high correlation.

\subsection{Approach}
Topics convey text meaning on a global level \citep{holtgraves1999comprehending}. In order to capture the deeper structure of meaning (advanced semantics), we hypothesized that calculating the distribution of document-topic probabilities from latent dirichlet allocation (LDA) \citep{Blei:03} could be helpful. 

Moreover, domain/world knowledge can be accounted for in LDA-resulting measures since LDA can be trained on various data. As explored in \citet{qumsiyeh2011readaid}, it may seem sensible to use the count of discovered topics as the measure of required knowledge. However, such measures can be extremely sensitive to passage length. Along with the count of discovered topics, we develop three others that consider how these topics are distributed: semantic richness, clarity, and noise. 

\begin{figure}[]
    \centering
    \includegraphics[width=0.41\textwidth]{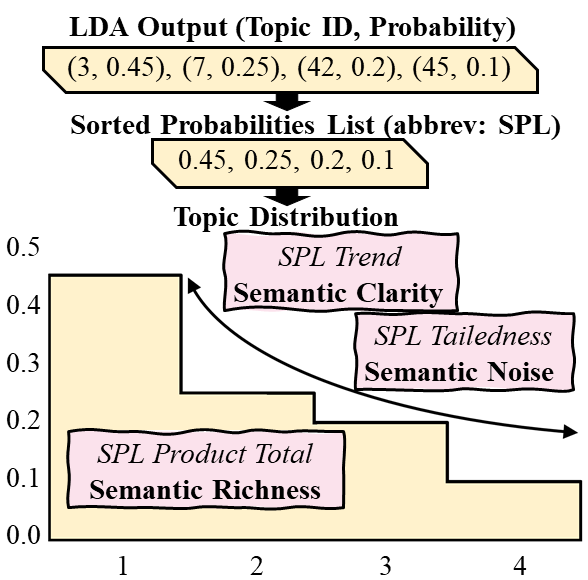}
    \caption{Graphical representation. Semantic Richness, Clarity, and Noise. abbrev: abbreviation.}
    \label{fig:1}
\end{figure}

Fig. 1 depicts the steps: 1. obtain output from a trained LDA model, 2. ignore topic ID and create a sorted probabilities list, and 3. calculate semantic richness, clarity, and noise. We model ``how'' the topics are distributed, not ``what'' the topics are.

\subsection{Semantic Richness}
Traditionally, semantic richness is quantified according to word usage \citep{pexman:08}. In a high-dimensional model of semantic space \citep{li2000acquisition}, co-occurring words clustered as semantic neighbors, quantifying semantic richness. As such, the previous models of semantic richness were often studied for word-level complexity and made no explicit connection with readability on a global level. Also, they were often subject-dependent \citep{buchanan2001characterizing}. As concluded by \citet{pexman:08}, semantic richness is defined in several ways. We propose a novel variation.

We apply the similar co-occurrence concept but on the passage level using LDA. Here, semantic richness is the measure of how ``largely'' populated the topics are. In fig. 1, we approximately define richness as the product total of SPL, which measures the count of discovered topics ($n$) and topic probability ($p$). Additionally, we multiply index ($i$) to reward longer $n$ so that the overall richness increases faster with more topics. See eqn. 1.

\begin{equation} \label{eq1}
\text{Semantic Richness} = \sum_{i=1}^{n}  p_i \cdot i
\end{equation}

\subsection{Semantic Clarity}
Semantic clarity is critical in understanding text \cite{peabody2016towards}. Likewise, complex meaning structures lead to comprehension difficulty \citep{pires2017towards}. Some existing studies quantify semantic complexity (or clarity) through various measures, but most on the fine line between lexical and semantic properties \citep{Collins-Thompson:14}. They rarely deal with the latent meaning representations or the clarity of the main topic.

For semantic clarity, we quantify how the probability distribution (fig. 1) is focused (skewed) towards the largest discovered topic. In other words, we hope to see how easily identifiable the main topic is. We wanted to adopt the standard skewness equation from statistics, but we developed an alternative (eqn. 2) because the standard equation failed to capture the anticipated trends in appendix A.

\begin{equation} \label{eq2}
\text{Semantic Clarity} = \frac{1}{n} \cdot \sum_{i=1}^{n}  (max(p) - p_{i})
\end{equation}

\subsection{Semantic Noise}
Semantic noise is the measure of the less-important topics (those with low probability), also the ``tailedness'' of sorted probability lists (fig. 1). A sorted probability list that resembles a (right-halved) leptokurtic curve would have higher semantic noise. In comparison, a (right-halved) platykurtic curve of similar length would have low semantic noise. We adopt the kurtosis equation under Fisher definition \citep{kokoska2000crc}. See eqn. 3.

\begin{equation} \label{eq3}
\text{Semantic Noise} = n \cdot \frac{\sum_{i=1}^{n} (p_i - \bar{p})^4}{(\sum_{i=1}^{n} (p_i - \bar{p})^2)^2}
\end{equation}

\section{Covered Features}
We study 255 linguistic features. For the already existing features, we add variations to widen coverage. The full list of features, feature codes, and definition are provided in appendix B. Also, we classify features into 14 subgroups. External depe-ndencies (e.g. parser) are reported in appendix D.

\subsection{Advanced Semantic Features (AdSem)}
Here, we follow the methods provided in section 2.

\textbf{1$\sim$3) Wikipedia (WoKF), WeeBit (WBKF), \& OneStop Knowledge Features (OSKF)}. 
Each subgroup name represents the respective training data. We train Online LDA \citep{hoffman2010online} with the 20210301 dump\footnote{dumps.wikimedia.org/enwiki} from English Wikipedia for WoKF. The others are trained on two popular corpora in RA: WeeBit and OneStopEnglish. 

For each training set, four variations of 50, 100, 150, 200 topics models are trained. Four features -- semantic richness, clarity, noise, and the total count of discovered topics -- are extracted per model.

\subsection{Discourse-Based Features (Disco)}
A text is more than a series of random sentences. It indicates a higher-level structure of dependencies.

\textbf{4) Entity Density Features (EnDF)}. 
Conceptual information is often introduced by entities. Hence, the count of entities affects the working memory burden \citep{Feng:09}. We bring entity-related features from \citet{Feng:10}.

\textbf{5) Entity Grid Features (EnGF)} 
Coherent texts are easy to comprehend. Thus, we measure coherence through entity grid, using the 16 transition pattern ratios approach by \citet{Pitler:08} as features. Also, we adopt local coherence scores \citep{guinaudeau2013graph}, using the code implemented by \citet{palma2018coherence}. 

\subsection{Syntactic Features (Synta)}
Syntactic complexity is associated with longer processing times \citep{gibson1998linguistic}. Such syntactic properties also affect the overall complexity of a text \citep{hale2016information}, an important indicator of readability.

\textbf{6) Phrasal Features (PhrF)}. 
Ratios involving clauses correlate with learners' abilities to read \citep{lu2010automatic}. We implement several variations, including the counts of noun, verb, and adverb phrases.

\textbf{7) Tree Structure Features (TrSF)}. 
We deal with the structural shape of parsed trees, inspired by the work on average parse tree height by \citet{Schwarm}. On a constituency parser (appendix D) output, NLTK \citep{loper2002nltk} is used for the final calculation of features.

\textbf{8) Part-of-Speech Features (POSF)}. 
Several studies report the effectiveness of using POS counts as features \citep{tonelli2012making, Leeb:20}. We count based on Universal POS tags\footnote{universaldependencies.org/u/pos}.

\subsection{Lexico-Semantic Features (LxSem)}
Perhaps the most explored, lexico-semantics capture the attributes associated with the difficulty or unfamiliarity of words \citep{Collins-Thompson:14}. 

\textbf{9) Variation Ratio Features (VarF)} 
\citet{Lu:11} reports noun, verb, adjective, and adverb variations, which represent the proportion of the respective category's words to total. We implement the features with variants from \citet{Vajjala:12}.

\textbf{10) Type Token Ratio Features (TTRF)}. 
TTR has been widely used as a measure of lexical richness in language acquisition studies \citep{Malvern:12}. We bring five variations of TTR from \citet{Vajjala:12}. For MTLD \citep{mccarthy2010mtld}, we default TTR to 0.72.

\textbf{11) Psycholinguistic Features (PsyF)} 
As explored in \citet{vajjala2016readability}, we implement various Age-of-Acquisition features from Kuperman study database \citet{kuperman2012age}.

\textbf{12) Word Familiarity Features (WorF)} 
Word frequency in a large representative corpus often represents lexical difficulty \citep{Collins-Thompson:14} due to unfamiliarity. We use SubtlexUS database \citep{brysbaert2009moving} to measure familiarity.

\subsection{Shallow Traditional Features (ShaTr)}
Classic readability formulas (e.g. Flesch-Kincaid Grade) \citep{Kincaid:75} or shallow measures often do not represent a specific linguistic branch.

\textbf{13) Shallow Features (ShaF)}
These features capture surface-level difficulty. Our measures include the average count of tokens and syllables.

\textbf{14) Traditional Formulas (TraF)}. 
For Flesh-Kincaid Grade Level, Automated Readability, and Gunning Fog, we follow the ``new'' formulas in \citet{Kincaid:75}. We follow \citet{si2001statistical} for Smog Index \citep{mc1969smog}. And we follow \citet{eltorai2015readability} for Linsear Write. 

\begin{figure*}[h]
    \centering
    \includegraphics[width=0.87\textwidth]{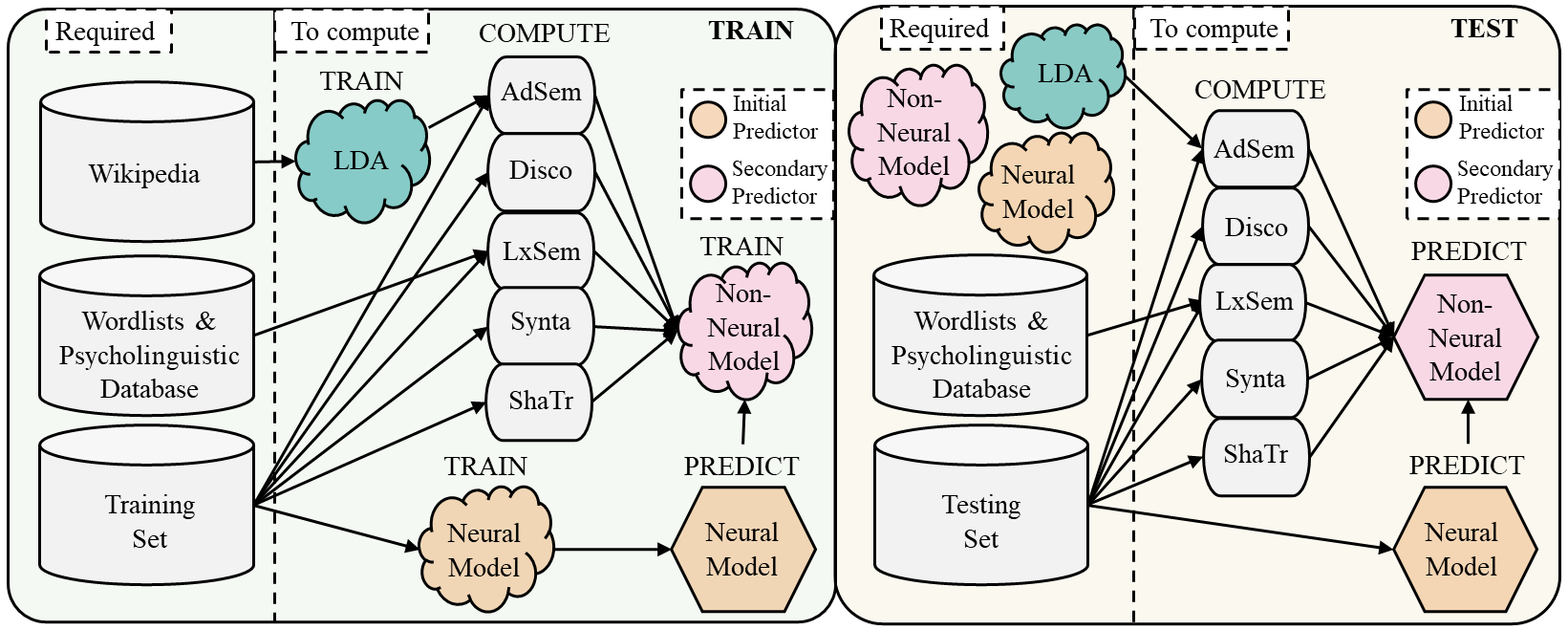}
    \caption{Hybrid model. AdSem, Disco, LxSem, Synta, and ShaTr show handcrafted features' linguistic branches.}
    \label{fig:2}
\end{figure*}

\section{Hybrid Model}
\subsection{Overview}
As shown in section 3, myriad linguistic properties affect readability. Despite the continual effort at handcrafting features, they lack coverage. \citet{deutsch-etal-2020-linguistic} hint neural models can better model the linguistic properties for RA task. But the performance/flexibility of neural models could improve. 

In our hybrid model, we take a simple approach of joining the soft label predictions of a neural model (e.g. BERT) with handcrafted features and wrapping it with a non-neural model (e.g. SVM). 

In fig. 2, the non-neural model (i.e. \textit{secondary predictor}) learns 1. predictions/outputs of the neural model and 2. handcrafted features. The addition of handcrafted features supplements what neural models (i.e. \textit{initial predictor}) might miss, reinforcing performance on the secondary prediction.

\subsection{Finding Best Combination}
Our hybrid architecture (fig. 2) is simple; \citet{deutsch-etal-2020-linguistic} explored a similar concept but did not achieve SOTA. But the benefits (section 4.1) from its simplicity are critical for RA, which has a lacking number/size of public datasets, wide educational use, and diverse handcrafted features. We obtain SOTA with a wider search on combinations. 

\subsubsection{Datasets and Evaluation Setups}
\begin{table}
\centering
\resizebox{0.47\textwidth}{!}{%
\begin{tabular}{c|ccc}
\hline\textbf{Properties}    & \textbf{WeeBit}   & \textbf{OneStopEng} & \textbf{Cambridge}        \\ 
\cmidrule(lr){1-1}\cmidrule(lr){2-2}\cmidrule(lr){3-3}\cmidrule(lr){4-4}
Target Audience               & General           & L2                & L2               \\
Covered Age                   & 7$\sim$16         & Adult             & A2$\sim$C2 (CEFR)          \\
Curriculum-Based?             & No                & No                & Yes             \\
Class-Balanced?               & No                & Yes               & No             \\
\# of Classes                 & 5                 & 3                 & 5                \\
\# of Items/Class             & 625               & 189               & 60               \\
\# of Tokens/Item             & 217               & 693               & 512              \\
Accessibility                 & Author            & Public            & Public           \\
\cmidrule(lr){1-4}
\end{tabular}
}
\caption{\label{Table 1} Statistics for datasets.}
\end{table}
\textbf{WeeBit}. Perhaps the most widely-used, WeeBit is often considered the gold standard in RA. It was first created as an expansion of the famous Weekly Reader corpus \citep{Feng:09}. To avoid classification bias, we downsample classes to equalize the number of items (passages) in each class to 625. It is common practice to downsample WeeBit.

\textbf{OneStopEnglish}. OneStopEnglish is an aligned passage corpus developed for RA and simplification studies. A passage is paraphrased into three readability classes. OneStopEnglish is designed to be a balanced dataset. No downsampling is needed.

\textbf{Cambridge}. Cambridge \citep{Xia:16} categorizes articles based on Cambridge English Exam levels (KET, PET, FCE, CAE, CPE). These five exams are targeted at learners at A2–C2 levels of the Common European Framework of Reference \citep{Xia:16}. We downsample to 60 items/class.

For evaluation, we calculate accuracy, weighted F1 score, precision, recall, and quadratic weighted kappa (QWK). The use of QWK is inspired by \citet{chen2016building,palma2019data}. We use stratified k-fold (k=5, train=0.8, val=0.1, test=0.1) and average the results for reliability. We use SciKit-learn \citep{pedregosa2011scikit} for metrics.

\subsubsection{Search on Neural Model}
\begin{table}[]
    \centering
    \resizebox{0.45\textwidth}{!}{%
    \begin{tabular}{cl|cccc}
    \hline 
    \multicolumn{2}{c|}{\textbf{Corpus}}      
                                 &\textbf{BERT} &\textbf{RoBERTa} &\textbf{BART}&\textbf{XLNet}  \\
    \cmidrule(lr){1-2}\cmidrule(lr){3-3}\cmidrule(lr){4-4}\cmidrule(lr){5-5}\cmidrule(lr){6-6}
    \multirow{5}{*}{WeeBit}      &Accuracy   &0.893            &\textbf{0.900}    &0.889                &0.881               \\
                                 &F1         &0.893            &\textbf{0.900}    &0.889                &0.880               \\
                                 &Precision  &0.896            &\textbf{0.902}    &0.892                &0.881               \\
                                 &Recall     &0.896            &\textbf{0.902}    &0.892                &0.881               \\
                                 &QWK        &0.966            &\textbf{0.970}    &0.963                &0.966               \\\cmidrule(lr){1-6}
    \multirow{5}{*}{OneStopE}    &Accuracy   &0.801            &0.965    &\textbf{0.968}                &0.804               \\
                                 &F1         &0.793            &0.965    &\textbf{0.968}                &0.794               \\
                                 &Precision  &0.815            &0.968    &\textbf{0.970}                &0.810               \\
                                 &Recall     &0.814            &0.968    &\textbf{0.970}                &0.810               \\
                                 &QWK        &0.840            &0.942    &\textbf{0.952}                &0.845               \\\cmidrule(lr){1-6}
    \multirow{5}{*}{Cambridge}   &Accuracy   &0.573            &\textbf{0.680}    &0.620                &0.573               \\
                                 &F1         &0.517            &\textbf{0.658}    &0.598                &0.554               \\
                                 &Precision  &0.528            &\textbf{0.693}    &0.643                &0.591               \\
                                 &Recall     &0.525            &\textbf{0.693}    &0.643                &0.591               \\
                                 &QWK        &0.809            &\textbf{0.881}    &0.835                &0.832               \\\cmidrule(lr){1-6}
    \cmidrule(lr){1-6}
    \end{tabular}
    }
\caption{\label{Append} Best performances, neural models.}
\end{table}
Extending from the existing use of BERT on RA \citep{deutsch-etal-2020-linguistic,Martinc:21}, we explore RoBERTa, \citep{liu2019roberta}, BART \citep{bart}, and XLNet \citep{XLNET:19}. We use base models for all (details in appendix D). For each of the four models (table 2), we perform grid searches on WeeBit validation sets to identify the well-performing hyperparameters based on 5-fold mean accuracy. Once identified, we used the same configuration for all the other corpora and performed no corpus-specific tweaking. We search the learning rates of [1e-5, 2e-5, 4e-5, 1e-4] and the batch sizes of [8, 16, 32]. The input sequence lengths are all set at 512, and we used Adam optimizer. Last, we fine-tuned the model for three epochs. Full hyperparameters are in appendix F.

In table 2, RoBERTa \& BART beat BERT \& XLNet on most metrics. \citet{Martinc:21} reports that transformers are weak on parallel datasets (OneStopEnglish) due to their reliance on semantic information. However, RoBERTa \& BART show great performances on OneStopEnglish as well. Such a phenomenon likely derives from numerous aspects of the architecture. We carefully posit that the varying pretraining steps could be a reason. 

BERT uses two objectives, masked language model (MLM) and next sentence prediction (NSP). The latter was included to capture the relation between sentences for natural language inference. Thus, sentence/segment-level input is used. Likewise, XLNet adopts a similar idea, limiting input to sentence/segment-level. But RoBERTa disproved the efficiency of NSP, adopting document-level inputs. Similarly, BART, via random shuffling of sentences and in-filling scheme, does not limit itself to a sentence/segment size input. As in section 3, ``readability'' is possibly a global-level representation (accumulated across the whole document). Thus, the performance differences could stem from the pretraining input size; sentence/segment-level input likely loses the global-level information.
\begin{table}[h]
    \centering
    \resizebox{0.45\textwidth}{!}{%
    \begin{tabular}{cl|cccc}
    \hline 
    \multicolumn{2}{c|}{\textbf{Corpus}}      &\textbf{SVM} &\textbf{RandomF} &\textbf{XGBoost}&\textbf{LogR}  \\
    \cmidrule(lr){1-2}\cmidrule(lr){3-3}\cmidrule(lr){4-4}\cmidrule(lr){5-5}\cmidrule(lr){6-6}
    \multirow{5}{*}{WeeBit}      &Accuracy   &\textbf{0.679} &0.638 &0.638          &0.622               \\
                                 &F1         &\textbf{0.672} &0.626 &0.627          &0.615               \\
                                 &Precision  &\textbf{0.696} &0.645 &0.656          &0.676               \\
                                 &Recall     &\textbf{0.679} &0.638 &0.638          &0.622               \\
                                 &QWK        &\textbf{0.716} &0.703 &0.692          &0.640               \\\cmidrule(lr){1-6}
    \multirow{5}{*}{OneStopE}    &Accuracy   &0.737          &0.709 &0.719          &\textbf{0.778}               \\
                                 &F1         &0.730          &0.706 &0.701          &\textbf{0.770}               \\
                                 &Precision  &0.751          &0.726 &0.734          &\textbf{0.778}               \\
                                 &Recall     &0.737          &0.709 &0.719          &\textbf{0.778}               \\
                                 &QWK        &0.400          &0.434 &0.363          &\textbf{0.486}     \\\cmidrule(lr){1-6}
    \multirow{5}{*}{Cambridge}   &Accuracy   &0.627          &0.673 &\textbf{0.685} &0.680               \\
                                 &F1         &0.613          &0.663 &\textbf{0.681} &0.657               \\
                                 &Precision  &0.660          &0.696 &\textbf{0.701} &0.694               \\
                                 &Recall     &0.627          &0.673 &\textbf{0.674} &0.680               \\
                                 &QWK        &0.857          &0.880 &\textbf{0.852} &0.855               \\
    \cmidrule(lr){1-6}
    \end{tabular}
    }
\caption{\label{Append} Best performances, non-neural models.}
\end{table}
\subsubsection{Search on Non-Neural Model}

We explored SVM, Random Forest (RandomF), Gradient Boosting (XGBoost) \citep{chen2016xgboost}, and Logistic Regression (LogR). With the exception of XGBoost, the chosen models are frequently used in RA but rarely go through adequate hyperparameter optimization steps \citep{ma2012ranking,yaneva2017combining,mohammadi2020machine}. We perform a randomized search to first identify the sensible range of hyperparameters to search. Then, we apply grid search to specify the optimal values. The parameters are in appendix F.

In table 3, we report the performances of the parameter-optimized models trained with all 255 handcrafted features. Compared to transformers, these non-neural models show lower accuracy in general. Even on the smallest Cambridge dataset, non-neural models do not necessarily show higher performances than transformers. But it is important to note that they managed to show fairly good, ``expectable'' performances on a much smaller dataset.

\begin{table}
\begin{subtable}{0.155\textwidth}
\centering
\resizebox{\textwidth}{!}{%
\begin{tabular}{c@{\hspace{0.8ex}}|c@{\hspace{0.8ex}}c@{\hspace{0.8ex}}}
\hline
\multirow{2.4}{*}{\textbf{Subgr}}& \multicolumn{2}{c}{\textbf{Model}} \\ 
\cmidrule(lr){2-3}
                          & LogR    & SVM           \\ 
\cmidrule(lr){1-1}\cmidrule(lr){2-2}\cmidrule(lr){3-3}
EnDF                      & 0.442   & 0.374   \\ 
ShaF                      & 0.404   & 0.409   \\
TrSF                      & 0.396   & 0.360   \\
POSF                      & 0.394   & 0.513   \\ 
\textbf{WorF}                      & 0.391   & 0.387   \\
\textbf{PsyF}                      & 0.378   & 0.437   \\
WoKF                      & 0.367   & 0.369   \\ \cmidrule(lr){1-3}
\end{tabular}
}
\caption{\textit{WeeBit} }\label{tabla4.1}
\end{subtable}
\begin{subtable}{0.155\textwidth}
\centering
\resizebox{\textwidth}{!}{%
\begin{tabular}{c@{\hspace{0.8ex}}|c@{\hspace{0.8ex}}c@{\hspace{0.8ex}}}
\hline
\multirow{2.4}{*}{\textbf{Subgr}}& \multicolumn{2}{c}{\textbf{Model}} \\ 
\cmidrule(lr){2-3}
                          & LogR    & SVM           \\ 
\cmidrule(lr){1-1}\cmidrule(lr){2-2}\cmidrule(lr){3-3}
TraF                      & 0.513   & 0.620   \\ 
\textbf{PsyF}                     & 0.437   & 0.696   \\
PhrF                      & 0.429   & 0.608   \\
VarF                      & 0.409   & 0.626   \\ 
TrSF                      & 0.391   & 0.614   \\
\textbf{WorF}                      & 0.387   & 0.637   \\
OSKF                      & 0.359   & 0.605   \\ \cmidrule(lr){1-3}
\end{tabular}
}
\caption{\textit{OneStopEnglish} }\label{tabla4.1}
\end{subtable}
\begin{subtable}{0.155\textwidth}
\centering
\resizebox{\textwidth}{!}{%
\begin{tabular}{c@{\hspace{0.8ex}}|c@{\hspace{0.8ex}}c@{\hspace{0.8ex}}}
\hline
\multirow{2.4}{*}{\textbf{Subgr}}& \multicolumn{2}{c}{\textbf{Model}} \\ 
\cmidrule(lr){2-3}
                          & LogR    & SVM           \\ 
\cmidrule(lr){1-1}\cmidrule(lr){2-2}\cmidrule(lr){3-3}
TraF                      & 0.640   & 0.593   \\ 
\textbf{WorF}             & 0.613   & 0.593   \\
ShaF                      & 0.600   & 0.587   \\
VarF                      & 0.600   & 0.533   \\ 
\textbf{PsyF}             & 0.593   & 0.620   \\
POSF                      & 0.553   & 0.407   \\
WoKF                      & 0.540   & 0.433   \\ \cmidrule(lr){1-3}
\end{tabular}
}
\caption{\textit{Cambridge} }\label{tabla4.1}
\end{subtable}
\caption{Top 7 Feature Subgroups.}
\end{table}
\begin{table*}[h]
    \begin{center}
    \resizebox{0.95\textwidth}{!}{
    \begin{tabular}{cl|@{\hspace{0.8ex}}c@{\hspace{0.8ex}}c@{\hspace{0.8ex}}c@{\hspace{0.8ex}}c@{\hspace{0.8ex}}|@{\hspace{0.5ex}}c@{\hspace{0.5ex}}c@{\hspace{0.8ex}}c@{\hspace{0.8ex}}|@{\hspace{0.8ex}}c@{\hspace{0.5ex}}c@{\hspace{0.8ex}}c@{\hspace{0.8ex}}|@{\hspace{0.8ex}}c@{\hspace{0.5ex}}c@{\hspace{0.8ex}}c@{\hspace{0.8ex}}|@{\hspace{0.8ex}}c@{\hspace{0.5ex}}c@{\hspace{0.8ex}}c@{\hspace{0.8ex}}}
    \hline 
    
    \multicolumn{2}{c|@{\hspace{0.8ex}}}{\multirow{5.2}{*}{
                                \textbf{Corpus}}} &\multicolumn{16}{c}{\textbf{Model}}  \\\cmidrule(lr){3-18}
                                &                 &\multicolumn{4}{c}{\textbf{Baselines, Previous Studies}} &\multicolumn{3}{c}{\textbf{BERT}} &\multicolumn{3}{c}{\textbf{RoBERTa}} &\multicolumn{3}{c}{\textbf{BART}} &\multicolumn{3}{c}{\textbf{XLNet}}\\  
                                 
                               \cmidrule(lr){3-6}\cmidrule(lr){7-9}\cmidrule(lr){10-12}\cmidrule(lr){13-15}\cmidrule(lr){16-18}  
    &                          &\textbf{Xia-16}&\textbf{Fili-19}&\multicolumn{2}{c|@{\hspace{0.8ex}}}{\textbf{Mar-21}}
                               &\textbf{hybrid}&$\Delta$&$\Delta$ 
                               &\textbf{hybrid}&$\Delta$&$\Delta$  
                               &\textbf{hybrid}&$\Delta$&$\Delta$ 
                               &\textbf{hybrid}&$\Delta$&$\Delta$ \\
                               
                               \cmidrule(lr){3-3}\cmidrule(lr){4-4}\cmidrule(lr){5-6}\cmidrule(lr){7-9}\cmidrule(lr){10-12}\cmidrule(lr){13-15}\cmidrule(lr){16-18}  
    &                          &\textbf{SVM}  &\textbf{LSTM}&\textbf{BERT}&\textbf{HAN}
                               &\textbf{GB-T1}&\textbf{BERT}&\textbf{GB}
                               &\textbf{RF-T1}&\textbf{RBRT}&\textbf{RF} 
                               &\textbf{RF-T1}&\textbf{BART}&\textbf{RF} 
                               &\textbf{RF-P3}&\textbf{XLNet}&\textbf{RF}\\ 
                               \cmidrule(lr){1-2} \cmidrule(lr){3-3}\cmidrule(lr){4-4}\cmidrule(lr){5-5}\cmidrule(lr){6-6}\cmidrule(lr){7-9}\cmidrule(lr){10-12}\cmidrule(lr){13-15}\cmidrule(lr){16-18}  
    \multirow{5}{*}{WeeBit}     &Accuracy  &0.803           &0.813&\textbf{0.857}&0.752          &0.895&0.002&0.257 &0.902         &0.002&0.264 &\textbf{0.905}&0.016&0.267 &0.892&0.011&0.254\\
                                &F1        &-               &-    &\textbf{0.866}&0.753          &0.895&0.002&0.268 &0.902         &0.002&0.276 &\textbf{0.905}&0.016&0.279 &0.892&0.012&0.266\\
                                &Precision &-               &-    &\textbf{0.857}&0.752          &0.897&0.001&0.241 &0.903         &0.001&0.258 &\textbf{0.905}&0.013&0.260 &0.893&0.012&0.248\\
                                &Recall    &-               &-    &\textbf{0.858}&0.752          &0.897&0.001&0.259 &0.903         &0.001&0.265 &\textbf{0.904}&0.012&0.266 &0.892&0.011&0.254\\
                                &QWK       &-               &-    &\textbf{0.953}&0.886          &0.969&0.001&0.277 &0.971         &0.001&0.268 &\textbf{0.968}&0.005&0.265 &0.966&0.000&0.263\\\cmidrule(lr){1-18}
    \multirow{5}{*}{OneStopE}   &Accuracy  &-               &-    &0.674         &\textbf{0.787} &0.982&0.181&0.263 &\textbf{0.990}&0.025&0.281 &0.971         &0.003&0.262 &0.848&0.044&0.139\\
                                &F1        &-               &-    &0.740         &\textbf{0.798} &0.982&0.189&0.281 &\textbf{0.995}&0.030&0.289 &0.971         &0.003&0.265 &0.848&0.050&0.142\\
                                &Precision &-               &-    &0.674         &\textbf{0.787} &0.983&0.168&0.249 &\textbf{0.995}&0.027&0.269 &0.972         &0.002&0.246 &0.852&0.042&0.126\\
                                &Recall    &-               &-    &0.677         &\textbf{0.789} &0.982&0.168&0.263 &\textbf{0.996}&0.028&0.287 &0.971         &0.001&0.262 &0.848&0.038&0.139\\
                                &QWK       &-               &-    &0.708         &\textbf{0.825} &0.973&0.133&0.610 &\textbf{0.996}&0.054&0.562 &0.952         &0.000&0.518 &0.855&0.010&0.369\\\cmidrule(lr){1-18}
    \multirow{5}{*}{Cambridge}  &Accuracy  &\textbf{$0.786^{**}$}&-    &-             &-              &0.687&0.114&0.002 &\textbf{0.763}&0.083&0.090 &0.727         &0.107&0.054 &0.687&0.114&0.014\\
                                &F1        &-               &-    &-             &-              &0.682&0.165&0.001 &\textbf{0.752}&0.094&0.089 &0.727         &0.129&0.064 &0.676&0.122&0.013\\
                                &Precision &-               &-    &-             &-              &0.732&0.204&0.031 &\textbf{0.792}&0.099&0.096 &0.760         &0.117&0.064 &0.710&0.119&0.014\\
                                &Recall    &-               &-    &-             &-              &0.687&0.162&0.013 &\textbf{0.753}&0.060&0.080 &0.727         &0.084&0.054 &0.687&0.096&0.014\\
                                &QWK       &-               &-    &-             &-              &0.873&0.064&0.021 &\textbf{0.919}&0.038&0.039 &0.889         &0.054&0.009 &0.888&0.056&0.008\\\cmidrule(lr){1-18}
    \end{tabular}
    }
    \end{center}
\begin{tablenotes}[para,flushleft]
\small
\item[**]\textit{Xia-16 (Cambridge) uses semi-supervised learning (self-training) on a larger corpus to increase performance.}
\end{tablenotes}
\caption{\label{Append} Best performances, hybrid models.}
\end{table*}

\subsubsection{Search on Handcrafted Features}
We start by ranking performances of the feature subgroups. In table 4, we report the top 7 (upper half) by accuracy on WeeBit. The result is obtained after training the respective model using the specified feature subgroup. Importantly, the advanced semantic features show good performance in all measures. WorF and PsyF, features calculated from external databases, rank in the top 7 for all corpora, hinting they are strong measures of readability. 

Moving on, we constructed several types of feature combinations with varying aims. These incl-ude: 1. \textbf{T-type} to thoroughly capture linguistic properties and 2. \textbf{P-type} to collect features by performance. We tested the variations on LogR and SVM to determine the optimal. Two sets (table 6) performed well. Appendix G reports all tested variations. We highlight that both advanced semantics and discourse added distinct (orthogonal) informa-tion, which was evidenced by performance change.

\subsection{Assembling Hybrid Model}

\begin{table}
\resizebox{0.47\textwidth}{!}{%
\begin{tabular}{@{\hspace{0.7ex}}c@{\hspace{0.7ex}}|@{\hspace{0.7ex}}l@{\hspace{0.7ex}}|@{\hspace{0.7ex}}c@{\hspace{0.7ex}}}
\hline
\textbf{Set} & \textbf{Features}                      & \textbf{LogR} \\ 
\cmidrule(lr){1-1}\cmidrule(lr){2-2}\cmidrule(lr){3-3}
T1  & AdSem+Disco+Synta+LxSem+ShaTr  & 0.622                    \\\cmidrule(lr){1-3}
P3  & ShaTr+EnDF+TrSF+POSF+WorF+PsyF+TraF+VarF  & 0.647                    \\\cmidrule(lr){1-3}
\end{tabular}
}
\begin{tablenotes}[para,flushleft]
\small
\item[*]\textit{Note: 5 letters (e.g. AdSem) mean linguistic branch. 4 letters (e.g. PhrF) mean subgroup. We report accuracy on WeeBit.}
\end{tablenotes}
\caption{\label{Table 1} Best feature sets.}
\end{table}
Based on the exploration so far, we assemble our hybrid model. We perform a brute-force grid search on four neural models (table 2), four non-neural models (table 3), and 14 feature sets (table 24). 

To interweave the model, we followed the steps of 1: obtain soft labels (probabilities that a text belongs to the respective readability class) from a neural model by softmax layer, 2: append the soft labels to handcrafted features (create a dataframe), 3. train non-neural model on the dataframe. As in fig 2, the neural models performed a sort of re-prediction to the data used for training to match the dataframe dimensions in training and test stages. 

Table 5 reports the best performing combination per respective neural model. Under ``hybrid'' column are code names (e.g. GB-T1 under BERT = XGBoost trained with handcrafted feature set T1 and BERT outputs). Under ``$\Delta$'' column, we report differences with the respective single model performance. We also include SOTA baseline results Xia-16 $\rightarrow$ \citet{Xia:16}, Fili-19 $\rightarrow$ \citet{Fili:19}, Mar-21 $\rightarrow$ \citet{Martinc:21}.

\subsection{Hybrid Model Results}
In table 5, our hybrid models achieve SOTA performances on WeeBit (BART-RF-T1) and OneSt-opEnglish (RoBERTa-RF-T1). With the exception of \citet{Xia:16} which uses extra data to increase accuracy, we also achieve SOTA on Cambridge: 76.3\% accuracy on a small dataset of only 60 items/class. Among the hybrids, RoBERTa-RF-T1 showed consistently high performance on all metrics. But all hybrid models beat previous SOTA results by a large margin. Notably, we achieve the near-perfect accuracy of 99\% on OneStopEnglish, a massive 20.3\% increase from the previous SOTA \citep{Martinc:21} by HAN \citep{10.1007/978-3-030-45439-5_3}.

Both neural and non-neural models benefit from the hybrid architecture. This is explicitly shown in BERT-GB-T1 performance on OneStopEnglish, achieving 98.2\% accuracy. This is an 18.1\% increase from BERT and a 26.3\% increase from XGBoost. However, BART did not benefit much from the hybrid architecture on WeeBit and OneStopEnglish, meaning that hybrid architectures do not augment model performance at all conditions. 

Along similar lines, the hybrid architecture performance on the larger WeeBit dataset showed only a small improvement from the transformer-only result. On the other hand, the hybrid architecture performance on the smaller Cambridge dataset was consistently better than the transformer-only performance. The hybrid shows $\sim$10\% improvement in accuracy on average for Cambridge. On the smallest dataset (Cambridge), the hybrid architecture benefited more from a non-neural, handcrafted features-based model like RF (Random Forest) and GB (XGBoost). On the largest dataset (WeeBit), the hybrid benefited more from a transformer. 

Our explanation is that the handcrafted features do not add much, at the data size of WeeBit. But the handcrafted features could be a great help where data is insufficient like they did for the Cambridge dataset. OneStopEnglish, being the medium-sized parallel dataset, could have hit the sweet spot for the hybrid architecture. But it must be noted that the data size is not the only determining factor as to which model (neural or non-neural) the hybrid architecture benefits more from. It must also be questioned if the max performance ($\because$ label noise induced by subjectivity) \citep{frenay2014comprehensive} is already achieved on WeeBit \citep{deutsch-etal-2020-linguistic}. 

Also, it seems that the hybrid architecture benefits when each model (neural and non-neural) already shows considerably good performance. This is plausible as the neural model outputs are considered features for the non-neural model. Including more ``fairly'' well-performing features only creates extra distractions. The hybrid architecture's limit is that it gets a model from ``good'' to ``great,'' not ``fair'' to ``good.'' But determining the definition of ``fair'' performance is a difficult feat as it likely depends on the dataset and a researcher's intuition from the empirical experience of the model. Hence, the hybrid architecture's limit is that one must test several combinations to pick the effective one.

\subsection{Why Not Directly Append Features?}
Regarding the model architecture, we examined appending the handcrafted features to transformer embeddings without the use of a secondary predictor like SVM. But an existing example of ReadNet \citep{10.1007/978-3-030-45439-5_3} hints that such a model is not robust to small datasets. ReadNet reports 52.8\% accuracy on Cambridge, worse than \textit{any} of our tested models (table 2, 3, 5). Besides, ReadNet claims to have achieved 91.7\% accuracy on WeeBit, without reports on downsampling. Many studies, like \citet{deutsch-etal-2020-linguistic}, report that the model accuracy can increase $\sim$4\% on the full, class-imbalanced WeeBit. Hence, ReadNet is not directly comparable. We omitted ReadNet from table 5.

\subsection{Why Was Our BERT Better?}
Noticeable in table 2 and table 5 is that our BERT implementation performed much better on WeeBit than what was reported. The dataset preparation methods and overall evaluation settings are the same or very similar across ours (accuracy: 89.3\%), \citet{deutsch-etal-2020-linguistic}'s (accuracy: 83.9\%), and \citet{Martinc:21}'s (accuracy: 85.7\%). We believe that the differences stem from the hyperparameters. 

Notably, \citet{deutsch-etal-2020-linguistic} uses 128 input sequence length. This is ineffective as the downsampled WeeBit has 2374 articles of over 128 tokens but only 275 articles of over 512 tokens (which was our input sequence length). Hence, we can reasonably think that much semantic information was lost in \citet{deutsch-etal-2020-linguistic}'s implementation. \citet{Martinc:21} uses 512 input sequence length but lacks a report on other possibly critical hyperparameters, and we cannot compare in detail.

\subsection{Data Size Effect}
\begin{figure}
    \raggedright
    \includegraphics[width=0.48\textwidth]{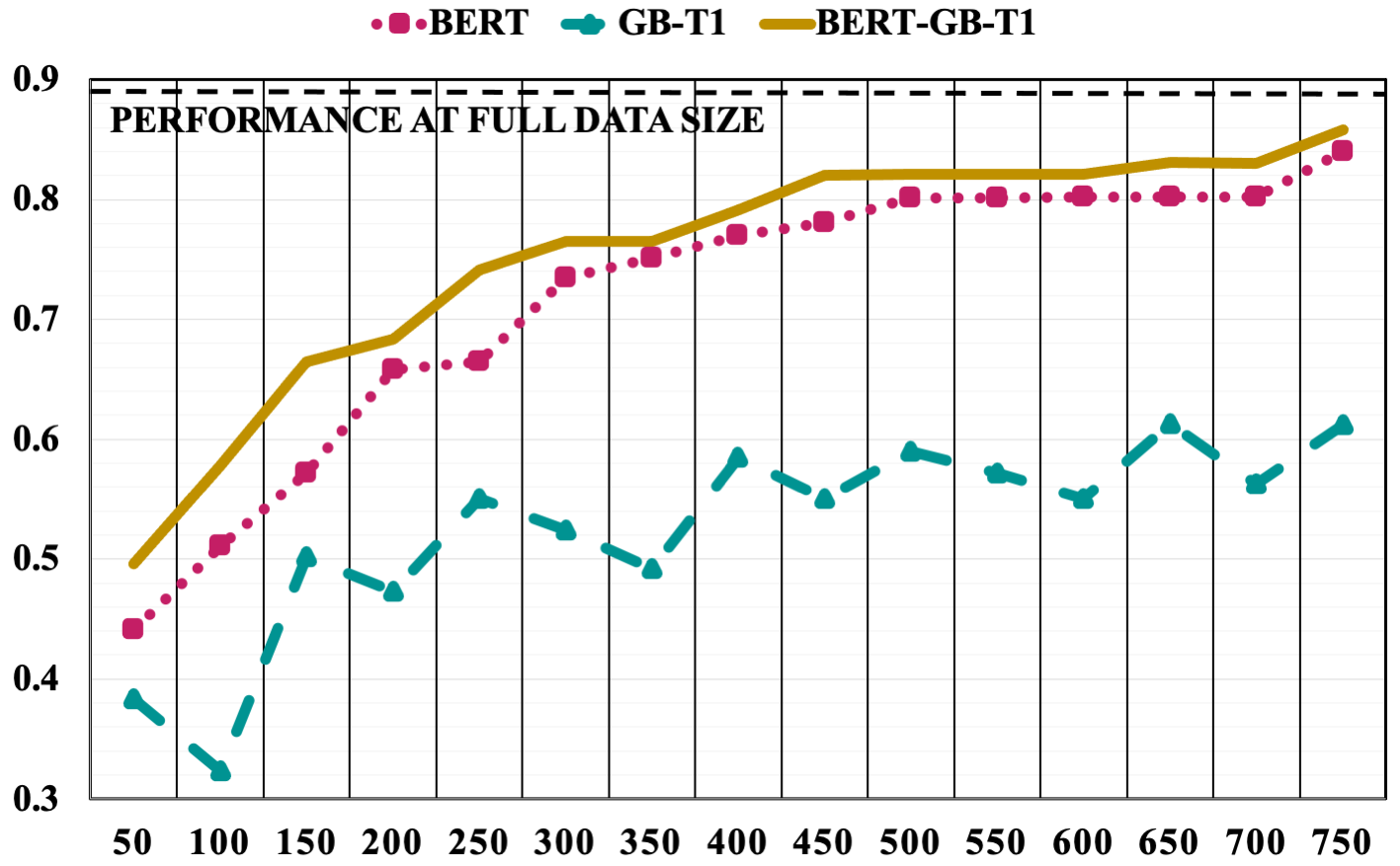}
    \caption{Performance Change, WeeBit Data Size}
    \label{fig:2}
\end{figure}
In table 5, our hybrid architecture generally did not contribute much to the classification on WeeBit. But we argue that it has much to do with data size.

To model how data size affects the accuracies of 1. hybrid model, 2. transformer, and 3. traditional ML, we conducted an additional experiment using the same test data (10\% of WeeBit) explained in section 4.2.1. However, we random sampled the train data (80\% of WeeBit) into the smaller sizes of from 50 to 750, with 50 passages increase each set. We sampled with equal class weights, meaning that a 250 passages train set has 50 from each readability class. We trained BERT-GB-T1 (table 5) on the sampled data and evaluated on the same test data throughout. We also recorded BERT and XGBoost (with T1 features) performances in fig. 3.

In fig. 3, the hybrid model performs consistently better than transformer (+0.01 $\sim$ 0.05) at all sizes. But the difference gap gets smaller as the train data size increases. The hybrid model does help the efficiency of learning RA linguistic properties. 

Contrary to the conventional beliefs, the transformer (BERT) performed better than our expectations, even on smaller data sizes. BERT always outperformed XGBoost. The traditional ML performance was arguably more consistent but never better than a transfomer's.

BERT-GB-T1's trend line seemed like the mixture of GB-T1's and BERT's. Notably, BERT-GB-T1 achieves 85.8\% accuracy on WeeBit using only 750 passages, 30\% of the original train data. For comparison, 85.7\% was the past SOTA (table 5).

\section{Cross Domain Evaluation}
99\% accuracy on OneStopEnglish (table 5) shows that our model is capable of almost perfectly capturing the linguistic properties relating to readability on certain datasets. This is a positive and abnormally quick improvement, considering that the reported RA accuracies have never exceeded 90\% on popular datasets \citep{Vajjala:12, xu2015problems, Xia:16, Vajjala:18} until 2021. Since the reported in-domain accuracies in RA had much room for improvement, we were not at the stage to be seriously concerned about cross-domain evaluation \citep{vstajner2018detailed} in this paper. 

It would be very favorable to run an extra cross-domain evaluation (which we believe to be a next-level topic). But realistically, performing a cross-domain evaluation requires a thorough study on at least two datasets, which is potentially out of scope in this research. The readability classes/levels are labeled by a few human experts, making the standards vary among datasets. To make two datasets suitable for cross-domain evaluation, much effort is needed to connect the two, such as the class mapping used in \citet{Xia:16}. However, it should be noted for future researchers that the notion of domain overfitting is indeed a common problem faced in RA, which often uses one dataset for train/test/validation. Without a new methodology to connect several datasets or a new large public dataset for RA, it will forever be challenging to develop a RA model for general use \citep{vajjala2021trends}. 

\section{Conclusion}
We have reported the four contributions mentioned in section 1. We checked that the new advanced semantic features add orthogonal information to the model. Further, we created hybrid models (table 5) that achieved SOTA results. RoBERTA-RF-T1 achieved 99\% accuracy on OneStopEnglish, and BERT-GB-T1 beat the previous SOTA on WeeBit using only 30\% of the original train data.

\section{Acknowledgements}
We wish to thank Dr. Inhwan Lee, Dongjun Lee, Sangjo Park, Donghyun Lee, and Eunsoo Shin. Partly funded by the 4th Industrial Revolution R\&D Program, Min. of SMEs \& Startups, Rep. of Korea.

\bibliography{anthology,custom}
\bibliographystyle{acl_natbib}

\appendix
\section{Trend, Advanced Semantic Features}
\begin{table}[H]
\centering
\resizebox{0.5\textwidth}{!}{%
\begin{tabular}{l@{\hspace{0.7ex}}|@{\hspace{0.5ex}}c@{\hspace{0.7ex}}c@{\hspace{0.5ex}}|@{\hspace{0.5ex}}c@{\hspace{0.7ex}}c@{\hspace{0.5ex}}|@{\hspace{0.5ex}}c@{\hspace{0.7ex}}c@{\hspace{0.5ex}}}
\hline \textbf{Sorted Probability List}&\textbf{R.}&\textbf{out}&\textbf{C.}&\textbf{out}&\textbf{N.}&\textbf{out}   \\\hline
9, 0.5, 0.5                            &Low        &115        &High       &56.7        &H-M        &30.0           \\
6, 2, 1, 0.5, 0.3, 0.2                 &L-M        &177        &H-M        &43.3        &High       &48.1           \\
4, 4, 1, 1                             &Mid        &190        &L-M        &15.0        &L-M        &18.5           \\
4, 2, 1, 1, 0.6, 0.4                   &H-M        &204        &Mid        &25.0        &Mid        &35.3           \\
2.5, 1.5, 1.5, 1.5, 1.5, 1.5           &High       &325        &Low        &8.34        &Low        &13.3           \\
\hline
\end{tabular}
}
\caption{\label{Append} Trends. Richness, Clarity, Noise. All numbers $\times 10$ for conciseness. L-M: Low-Mid. H-M: High-Mid.}
\end{table}

In table 7, we name each list as 1 $\sim$ 5 from top to bottom. ``out'' refers to raw output from equations. See what the sorted probabilities list is in fig. 1.

\textbf{Semantic Richness}. List 4  and list 5 have the same lengths. However, list 5 contains more meaningful topics ($\uparrow p$) throughout the list, resulting in higher overall semantic richness. As such, semantic richness rewards long probability lists ($\uparrow n$) with more meaningful ($\uparrow p$) topics. Similarly, list 3 ($\downarrow n$,$\uparrow p$) has higher richness than list 2 ($\uparrow n$,$\downarrow p$).

\textbf{Semantic Clarity}. List 3 and list 4 have the same $max(p)$ and two other same elements ($1$). However, the second element in list 3 is the same as the first element, resulting in increased difficulty in identifying the main topic ($max(p)$). Likewise, semantic clarity rewards the deviation between the $max(p)$ and the other elements \& short probability lists ($\downarrow n$). Hence, list 1 has the highest clarity.

\textbf{Semantic Noise}. List 2 and list 4 have the same lengths of 6 elements. However, list 2 contains more extraneous topics ($\downarrow p$), resulting in higher semantic noise. As such, semantic noise rewards longer lists ($\uparrow n$) with more extraneous elements ($\downarrow p$). As a result, list 5 has the least semantic noise.

\section{Features, Codes, and Definitions}
\begin{table}[H]
    \centering
    \resizebox{0.5\textwidth}{!}{%
    \begin{tabular}{l@{\hspace{0.3ex}}|l@{\hspace{0.5ex}}|l@{\hspace{0.5ex}}}
    \hline
        \textbf{$idx$} & \textbf{Code} & \textbf{Definition} \\ \hline
        1              & WRich05\_S     &Richness, 50 topics extracted from Wikipedia Dump        \\
        2              & WClar05\_S     &Clarity, 50 topics extracted from Wikipedia Dump         \\
        3              & WNois05\_S     &Noise, 50 topics extracted from Wikipedia Dump           \\
        4              & WTopc05\_S     &\# of topics, 50 topics extracted from Wikipedia Dump        \\
        5              & WRich10\_S     &Richness, 100 topics extracted from Wikipedia Dump      \\
        6              & WClar10\_S     &Clarity, 100 topics extracted from Wikipedia Dump       \\
        7              & WNois10\_S     &Noise, 100 topics extracted from Wikipedia Dump         \\
        8              & WTopc10\_S     &\# of topics, 100 topics extracted from Wikipedia Dump       \\
        9              & WRich15\_S     &Richness, 150 topics extracted from Wikipedia Dump      \\
        10             & WClar15\_S     &Clarity, 150 topics extracted from Wikipedia Dump       \\
        11             & WNois15\_S     &Noise, 150 topics extracted from Wikipedia Dump         \\
        12             & WTopc15\_S     &\# of topics, 150 topics extracted from Wikipedia Dump       \\
        13             & WRich20\_S     &Richness, 200 topics extracted from Wikipedia Dump      \\
        14             & WClar20\_S     &Clarity, 200 topics extracted from Wikipedia Dump       \\
        15             & WNois20\_S     &Noise, 200 topics extracted from Wikipedia Dump         \\
        16             & WTopc20\_S     &\# of topics, 200 topics extracted from Wikipedia Dump       \\    \hline  
    \end{tabular}
    }
\caption{\label{Append} Wikipedia Knowledge Features (WoKF).}
\end{table}
\begin{table}[H]
    \centering
    \resizebox{0.5\textwidth}{!}{%
    \begin{tabular}{l@{\hspace{0.3ex}}|l@{\hspace{0.5ex}}|l@{\hspace{0.5ex}}}
    \hline
        \textbf{$idx$} & \textbf{Code} & \textbf{Definition} \\ \hline
        17             & BRich05\_S     &Richness, 50 topics extracted from WeeBit Corpus      \\        
        18             & BClar05\_S     &Clarity, 50 topics extracted from WeeBit Corpus       \\         
        19             & BNois05\_S     &Noise, 50 topics extracted from WeeBit Corpus         \\         
        20             & BTopc05\_S     &\# of topics, 50 topics extracted from WeeBit Corpus       \\         
        21             & BRich10\_S     &Richness, 100 topics extracted from WeeBit Corpus     \\         
        22             & BClar10\_S     &Clarity, 100 topics extracted from WeeBit Corpus      \\        
        23             & BNois10\_S     &Noise, 100 topics extracted from WeeBit Corpus        \\        
        24             & BTopc10\_S     &\# of topics, 100 topics extracted from WeeBit Corpus      \\        
        25             & BRich15\_S     &Richness, 150 topics extracted from WeeBit Corpus     \\         
        26             & BClar15\_S     &Clarity, 150 topics extracted from WeeBit Corpus      \\        
        27             & BNois15\_S     &Noise, 150 topics extracted from WeeBit Corpus        \\        
        28             & BTopc15\_S     &\# of topics, 150 topics extracted from WeeBit Corpus      \\        
        29             & BRich20\_S     &Richness, 200 topics extracted from WeeBit Corpus     \\         
        30             & BClar20\_S     &Clarity, 200 topics extracted from WeeBit Corpus      \\         
        31             & BNois20\_S     &Noise, 200 topics extracted from WeeBit Corpus        \\         
        32             & BTopc20\_S     &\# of topics, 200 topics extracted from WeeBit Corpus      \\ \hline         
    \end{tabular}
    }
\caption{\label{Append} WeeBit Knowledge Features (WBKF).}
\end{table}
\begin{table}[H]
    \centering
    \resizebox{0.5\textwidth}{!}{%
    \begin{tabular}{l@{\hspace{0.3ex}}|l@{\hspace{0.5ex}}|l@{\hspace{0.5ex}}}
    \hline
        \textbf{$idx$} & \textbf{Code} & \textbf{Definition} \\ \hline
        33             & ORich05\_S     &Richness, 50 topics extracted from OneStop Corpus    \\          
        34             & OClar05\_S     &Clarity, 50 topics extracted from OneStop Corpus     \\           
        35             & ONois05\_S     &Noise, 50 topics extracted from OneStop Corpus       \\           
        36             & OTopc05\_S     &\# of topics, 50 topics extracted from OneStop Corpus     \\           
        37             & ORich10\_S     &Richness, 100 topics extracted from OneStop Corpus   \\          
        38             & OClar10\_S     &Clarity, 100 topics extracted from OneStop Corpus    \\          
        39             & ONois10\_S     &Noise, 100 topics extracted from OneStop Corpus      \\          
        40             & OTopc10\_S     &\# of topics, 100 topics extracted from OneStop Corpus    \\          
        41             & ORich15\_S     &Richness, 150 topics extracted from OneStop Corpus   \\          
        42             & OClar15\_S     &Clarity, 150 topics extracted from OneStop Corpus    \\          
        43             & ONois15\_S     &Noise, 150 topics extracted from OneStop Corpus      \\          
        44             & OTopc15\_S     &\# of topics, 150 topics extracted from OneStop Corpus    \\          
        45             & ORich20\_S     &Richness, 200 topics extracted from OneStop Corpus   \\          
        46             & OClar20\_S     &Clarity, 200 topics extracted from OneStop Corpus    \\           
        47             & ONois20\_S     &Noise, 200 topics extracted from OneStop Corpus      \\           
        48             & OTopc20\_S     &\# of topics, 200 topics extracted from OneStop Corpus    \\ \hline                
    \end{tabular}
    }
\caption{\label{Append} OneStop Knowledge Features (OSKF).}
\end{table}
\begin{table}[H]
    \centering
    \resizebox{0.5\textwidth}{!}{%
    \begin{tabular}{l@{\hspace{0.3ex}}|l@{\hspace{0.5ex}}|l@{\hspace{0.5ex}}}
    \hline
        \textbf{$idx$} & \textbf{Code} & \textbf{Definition} \\ \hline
        49             & to\_EntiM\_C    & total number of Entities Mentions                       \\              
        50             & as\_EntiM\_C    & average number of Entities Mentions per sentence        \\              
        51             & at\_EntiM\_C    & average number of Entities Mentions per token (word)           \\              
        52             & to\_UEnti\_C    & total number of unique Entities                                \\              
        53             & as\_UEnti\_C    & average number of unique Entities per sentence                 \\              
        54             & at\_UEnti\_C    & average number of unique Entities per token (word)             \\ \hline                           
    \end{tabular}
    }
\caption{\label{Append} Entity Density Features (EnDF).}
\end{table}
\begin{table}[H]
    \centering
    \resizebox{0.5\textwidth}{!}{%
    \begin{tabular}{l@{\hspace{0.3ex}}|l@{\hspace{0.7ex}}|l@{\hspace{0.ex}}}
    \hline
        \textbf{$idx$} & \textbf{Code}  & \textbf{Definition} \\ \hline
        55             & ra\_SSToT\_C   & ratio of SS transitions $:$ total, count from Entity Grid         \\
        56             & ra\_SOToT\_C   & ratio of SO transitions $:$ total, count from Entity Grid         \\
        57             & ra\_SXToT\_C   & ratio of SX transitions $:$ total, count from Entity Grid         \\
        58             & ra\_SNToT\_C   & ratio of SN transitions $:$ total, count from Entity Grid         \\
        59             & ra\_OSToT\_C   & ratio of OS transitions $:$ total, count from Entity Grid         \\
        60             & ra\_OOToT\_C   & ratio of OO transitions $:$ total, count from Entity Grid         \\
        61             & ra\_OXToT\_C   & ratio of OX transitions $:$ total, count from Entity Grid         \\
        62             & ra\_ONToT\_C   & ratio of ON transitions $:$ total, count from Entity Grid         \\
        63             & ra\_XSToT\_C   & ratio of XS transitions $:$ total, count from Entity Grid         \\
        64             & ra\_XOToT\_C   & ratio of XO transitions $:$ total, count from Entity Grid         \\
        65             & ra\_XXToT\_C   & ratio of XX transitions $:$ total, count from Entity Grid         \\
        66             & ra\_XNToT\_C   & ratio of XN transitions $:$ total, count from Entity Grid         \\
        67             & ra\_NSToT\_C   & ratio of NS transitions $:$ total, count from Entity Grid         \\
        68             & ra\_NOToT\_C   & ratio of NO transitions $:$ total, count from Entity Grid         \\
        69             & ra\_NXToT\_C   & ratio of NX transitions $:$ total, count from Entity Grid         \\
        70             & ra\_NNToT\_C   & ratio of NN transitions $:$ total, count from Entity Grid         \\ \hline
        \end{tabular}
    }
\caption{\label{Append} Entity Grid Features (EnDF) Part 1.}
\end{table}
\begin{table}[H]
    \centering
    \resizebox{0.5\textwidth}{!}{%
    \begin{tabular}{l@{\hspace{0.3ex}}|l@{\hspace{0.5ex}}|l@{\hspace{0.5ex}}}
    \hline
        \textbf{$idx$} & \textbf{Code}  & \textbf{Definition} \\ \hline
        71             & LoCohPA\_S     & Local Coherence for PA score from Entity Grid          \\
        72             & LoCohPW\_S     & Local Coherence for PW score from Entity Grid          \\
        73             & LoCohPU\_S     & Local Coherence for PU score from Entity Grid          \\
        74             & LoCoDPA\_S     & Local Coherence dist. for PA score from Entity Grid \\
        75             & LoCoDPW\_S     & Local Coherence dist. for PW score from Entity Grid \\
        76             & LoCoDPU\_S     & Local Coherence dist. for PU score from Entity Grid \\ \hline                      
    \end{tabular}
    }
\caption{\label{Append} Entity Grid Features (EnDF) Part 2.}
\end{table}
\begin{table}[H]
    \centering
    \resizebox{0.5\textwidth}{!}{%
    \begin{tabular}{l@{\hspace{0.5ex}}|l@{\hspace{0.5ex}}|l@{\hspace{0.8ex}}}
    \hline
        \textbf{$idx$} & \textbf{Code}  & \textbf{Definition} \\ \hline
        77     & to\_NoPhr\_C   & total count of Noun phrases                                     \\
        78     & as\_NoPhr\_C   & average count of Noun phrases per sentence                      \\
        79     & at\_NoPhr\_C   & average count of Noun phrases per token                         \\
        80     & ra\_NoVeP\_C   & ratio of Noun phrases : Verb phrases count               \\
        81     & ra\_NoSuP\_C   & ratio of Noun phrases : Subordinate clauses count        \\
        82     & ra\_NoPrP\_C   & ratio of Noun phrases : Prep phrases count               \\
        83     & ra\_NoAjP\_C   & ratio of Noun phrases : Adj phrases count                \\
        84     & ra\_NoAvP\_C   & ratio of Noun phrases : Adv phrases count                \\
        85     & to\_VePhr\_C   & total count of Verb phrases                                     \\
        86     & as\_VePhr\_C   & average count of Verb phrases per sentence                      \\
        87     & at\_VePhr\_C   & average count of Verb phrases per token                         \\
        88     & ra\_VeNoP\_C   & ratio of Verb phrases : Noun phrases count               \\
        89     & ra\_VeSuP\_C   & ratio of Verb phrases : Subordinate clauses count        \\
        90     & ra\_VePrP\_C   & ratio of Verb phrases : Prep phrases count               \\
        91     & ra\_VeAjP\_C   & ratio of Verb phrases : Adj phrases count                \\
        92     & ra\_VeAvP\_C   & ratio of Verb phrases : Adv phrases count                \\
        93     & to\_SuPhr\_C   & total count of Subordinate clauses                              \\
        94     & as\_SuPhr\_C   & average count of Subordinate clauses per sentence               \\
        95     & at\_SuPhr\_C   & average count of Subordinate clauses per token                  \\
        96     & ra\_SuNoP\_C   & ratio of Subordinate clauses : Noun phrases count        \\
        97     & ra\_SuVeP\_C   & ratio of Subordinate clauses : Verb phrases count        \\
        98     & ra\_SuPrP\_C   & ratio of Subordinate clauses : Prep phrases count        \\
        99     & ra\_SuAjP\_C   & ratio of Subordinate clauses : Adj phrases count         \\
        100    & ra\_SuAvP\_C   & ratio of Subordinate clauses : Adv phrases count         \\
        101    & to\_PrPhr\_C   & total count of prepositional phrases                            \\
        102    & as\_PrPhr\_C   & average count of prepositional phrases per sentence             \\
        103    & at\_PrPhr\_C   & average count of prepositional phrases per token                \\
        104    & ra\_PrNoP\_C   & ratio of Prep phrases : Noun phrases count               \\
        105    & ra\_PrVeP\_C   & ratio of Prep phrases : Verb phrases count               \\
        106    & ra\_PrSuP\_C   & ratio of Prep phrases : Subordinate clauses count        \\
        107    & ra\_PrAjP\_C   & ratio of Prep phrases : Adj phrases count                \\
        108    & ra\_PrAvP\_C   & ratio of Prep phrases : Adv phrases count                \\
        109    & to\_AjPhr\_C   & total count of Adjective phrases                                \\
        110    & as\_AjPhr\_C   & average count of Adjective phrases per sentence                 \\
        111    & at\_AjPhr\_C   & average count of Adjective phrases per token                    \\
        112    & ra\_AjNoP\_C   & ratio of Adj phrases : Noun phrases count                \\
        113    & ra\_AjVeP\_C   & ratio of Adj phrases : Verb phrases count                \\
        114    & ra\_AjSuP\_C   & ratio of Adj phrases : Subordinate clauses count         \\
        115    & ra\_AjPrP\_C   & ratio of Adj phrases : Prep phrases count                \\
        116    & ra\_AjAvP\_C   & ratio of Adj phrases : Adv phrases count                 \\
        117    & to\_AvPhr\_C   & total count of Adverb phrases                                   \\
        118    & as\_AvPhr\_C   & average count of Adverb phrases per sentence                    \\
        119    & at\_AvPhr\_C   & average count of Adverb phrases per token                       \\
        120    & ra\_AvNoP\_C   & ratio of Adv phrases : Noun phrases count                \\
        121    & ra\_AvVeP\_C   & ratio of Adv phrases : Verb phrases count                \\
        122    & ra\_AvSuP\_C   & ratio of Adv phrases : Subordinate clauses count         \\
        123    & ra\_AvPrP\_C   & ratio of Adv phrases : Prep phrases count                \\
        124    & ra\_AvAjP\_C   & ratio of Adv phrases : Adj phrases count                 \\    \hline          
    \end{tabular}
    }
\caption{\label{Append} Phrasal Features (PhrF)}
\end{table}
\begin{table}[H]
    \centering
    \resizebox{0.5\textwidth}{!}{%
    \begin{tabular}{l@{\hspace{0.3ex}}|l@{\hspace{0.7ex}}|l@{\hspace{0.7ex}}}
    \hline
        \textbf{$idx$} & \textbf{Code}  & \textbf{Definition} \\ \hline
    125   & to\_TreeH\_C   & total parsed Tree Height of all sentences                              \\
    126   & as\_TreeH\_C   & average parsed Tree Height per sentence                                \\
    127   & at\_TreeH\_C   & average parsed Tree Height per token                           \\
    128   & to\_FTree\_C   & total length of Flattened parsed Trees                                 \\
    129   & as\_FTree\_C   & average length of Flattened parsed Trees per sentence                  \\
    130   & at\_FTree\_C   & average length of Flattened parsed Trees per token              \\     \hline                
    \end{tabular}
    }
\caption{\label{Append} Tree Structural Features (TrSF)}
\end{table}
\begin{table}[H]
    \centering
    \resizebox{0.5\textwidth}{!}{%
    \begin{tabular}{l@{\hspace{0.3ex}}|l@{\hspace{0.3ex}}|l@{\hspace{0.3ex}}}
    \hline
        \textbf{$idx$} & \textbf{Code}  & \textbf{Definition} \\ \hline
    131   & to\_NoTag\_C   & total count of Noun tags                                                   \\
    132   & as\_NoTag\_C   & average count of Noun tags per sentence                                    \\
    133   & at\_NoTag\_C   & average count of Noun tags per token                                       \\
    134   & ra\_NoAjT\_C   & ratio of Noun : Adjective count                                 \\
    135   & ra\_NoVeT\_C   & ratio of Noun : Verb count                                      \\
    136   & ra\_NoAvT\_C   & ratio of Noun : Adverb count                                    \\
    137   & ra\_NoSuT\_C   & ratio of Noun : Subordinating Conj. count                     \\
    138   & ra\_NoCoT\_C   & ratio of Noun : Coordinating Conj. count                      \\
    139   & to\_VeTag\_C   & total count of Verb tags                                                   \\
    140   & as\_VeTag\_C   & average count of Verb tags per sentence                                    \\
    141   & at\_VeTag\_C   & average count of Verb tags per token                                       \\
    142   & ra\_VeAjT\_C   & ratio of Verb : Adjective count                                 \\
    143   & ra\_VeNoT\_C   & ratio of Verb : Noun count                                      \\
    144   & ra\_VeAvT\_C   & ratio of Verb : Adverb count                                    \\
    145   & ra\_VeSuT\_C   & ratio of Verb : Subordinating Conj. count                     \\
    146   & ra\_VeCoT\_C   & ratio of Verb : Coordinating Conj. count                      \\
    147   & to\_AjTag\_C   & total count of Adjective tags                                              \\
    148   & as\_AjTag\_C   & average count of Adjective tags per sentence                               \\
    149   & at\_AjTag\_C   & average count of Adjective tags per token                                  \\
    150   & ra\_AjNoT\_C   & ratio of Adjective : Noun count                                 \\
    151   & ra\_AjVeT\_C   & ratio of Adjective : Verb count                                 \\
    152   & ra\_AjAvT\_C   & ratio of Adjective : Adverb count                               \\
    153   & ra\_AjSuT\_C   & ratio of Adjective : Subordinating Conj. count                \\
    154   & ra\_AjCoT\_C   & ratio of Adjective : Coordinating Conj. count                 \\
    155   & to\_AvTag\_C   & total count of Adverb tags                                                 \\
    156   & as\_AvTag\_C   & average count of Adverb tags per sentence                                  \\
    157   & at\_AvTag\_C   & average count of Adverb tags per token                                     \\
    158   & ra\_AvAjT\_C   & ratio of Adverb : Adjective count                               \\
    159   & ra\_AvNoT\_C   & ratio of Adverb : Noun count                                    \\
    160   & ra\_AvVeT\_C   & ratio of Adverb : Verb count                                    \\
    161   & ra\_AvSuT\_C   & ratio of Adverb : Subordinating Conj. count                   \\
    162   & ra\_AvCoT\_C   & ratio of Adverb : Coordinating Conj. count                    \\
    163   & to\_SuTag\_C   & total count of Subordinating Conj. tags                              \\
    164   & as\_SuTag\_C   & average count of Subordinating Conj. per sentence               \\
    165   & at\_SuTag\_C   & average count of Subordinating Conj. per token                  \\
    166   & ra\_SuAjT\_C   & ratio of Subordinating Conj. : Adjective count            \\
    167   & ra\_SuNoT\_C   & ratio of Subordinating Conj. : Noun count                 \\
    168   & ra\_SuVeT\_C   & ratio of Subordinating Conj. : Verb count                 \\
    169   & ra\_SuAvT\_C   & ratio of Subordinating Conj. : Adverb count               \\
    170   & ra\_SuCoT\_C   & ratio, Subordinating Conj. : Coordinating Conj. count \\
    171   & to\_CoTag\_C   & total count of Coordinating Conj. tags                               \\
    172   & as\_CoTag\_C   & average count of Coordinating Conj. per sentence                \\
    173   & at\_CoTag\_C   & average count of Coordinating Conj. per token                   \\
    174   & ra\_CoAjT\_C   & ratio of Coordinating Conj. : Adjective count             \\
    175   & ra\_CoNoT\_C   & ratio of Coordinating Conj. : Noun count                  \\
    176   & ra\_CoVeT\_C   & ratio of Coordinating Conj. : Verb count                  \\
    177   & ra\_CoAvT\_C   & ratio of Coordinating Conj. : Adverb count                \\
    178   & ra\_CoSuT\_C   & ratio, Coordinating Conj. : Subordinating Conj. count \\
    179   & to\_ContW\_C   & total count of Content words                           \\
    180   & as\_ContW\_C   & average count of Content words per sentence            \\
    181   & at\_ContW\_C   & average count of Content words per token               \\
    182   & to\_FuncW\_C   & total count of Function words                          \\
    183   & as\_FuncW\_C   & average count of Function words per sentence           \\
    184   & at\_FuncW\_C   & average count of Function words per token              \\
    185   & ra\_CoFuW\_C   & ratio of Content words to Function words               \\ \hline             
    \end{tabular}
    }
\caption{\label{Append} Part-of-Speech Features (POSF)}
\end{table}
\begin{table}[H]
    \centering
    \resizebox{0.5\textwidth}{!}{%
    \begin{tabular}{l@{\hspace{0.3ex}}|l@{\hspace{0.3ex}}|l@{\hspace{0.3ex}}}
    \hline
        \textbf{$idx$} & \textbf{Code}  & \textbf{Definition} \\ \hline
    186   & SimpNoV\_S    & unique Nouns/total Nouns \#Noun Variation                   \\
    187   & SquaNoV\_S    & (unique Nouns**2)/total Nouns \#Squared Noun Variation      \\
    188   & CorrNoV\_S    & unique Nouns/sqrt(2*total Nouns) \#Corrected Noun Var \\
    189   & SimpVeV\_S    & unique Verbs/total Verbs \#Verb Variation                   \\
    190   & SquaVeV\_S    & (unique Verbs**2)/total Verbs \#Squared Verb Variation      \\
    191   & CorrVeV\_S    & unique Verbs/sqrt(2*total Verbs) \#Corrected Verb Var \\
    192   & SimpAjV\_S    & unique Adjectives/total Adjectives \#Adjective Var    \\
    193   & SquaAjV\_S    & (unique Adj**2)/total Adj \#Squared Adj Variation       \\
    194   & CorrAjV\_S    & unique Adj/sqrt(2*total Adj) \#Corrected Adj Var  \\
    195   & SimpAvV\_S    & unique Adverbs/total Adverbs \#Adverb Variation             \\
    196   & SquaAvV\_S    & (unique Adv**2)/total Adv \#Squared Adv Variation           \\
    197   & CorrAvV\_S    & unique Adv/sqrt(2*total Adv) \#Corrected Adv Var      \\ \hline        
    \end{tabular}
    }
\caption{\label{Append} Variation Ratio Features (VarF)}
\end{table}
\begin{table}[H]
    \centering
    \resizebox{0.5\textwidth}{!}{%
    \begin{tabular}{l@{\hspace{0.3ex}}|l@{\hspace{0.3ex}}|l@{\hspace{0.3ex}}}
    \hline
        \textbf{$idx$} & \textbf{Code}  & \textbf{Definition} \\ \hline
    198   & SimpTTR\_S    & unique tokens/total tokens \#TTR                                     \\
    199   & CorrTTR\_S    & unique/sqrt(2*total) \#Corrected TTR                   \\
    200   & BiLoTTR\_S    & log(unique)/log(total) \#Bi-Logarithmic TTR            \\
    201   & UberTTR\_S    & (log(unique))$^2$/log(total/unique) \#Uber  \\
    202   & MTLDTTR\_S    & \#Measure of Textual Lexical Diversity (TTR, 0.72)            \\   \hline
    \end{tabular}
    }
\caption{\label{Append} Type Token Ratio Features (TTRF)}
\end{table}
\begin{table}[H]
    \centering
    \resizebox{0.5\textwidth}{!}{%
    \begin{tabular}{l@{\hspace{0.3ex}}|l@{\hspace{0.5ex}}|l@{\hspace{0.5ex}}}
    \hline
        \textbf{$idx$} & \textbf{Code}  & \textbf{Definition} \\ \hline
    203   & to\_AAKuW\_C   & total AoA (Age of Acquisition) of words, Kuperman                       \\
    204   & as\_AAKuW\_C   & average AoA of words per sentence, Kuperman                              \\
    205   & at\_AAKuW\_C   & average AoA of words per token, Kuperman                                 \\
    206   & to\_AAKuL\_C   & total AoA of lemmas, Kuperman                                     \\
    207   & as\_AAKuL\_C   & average AoA of lemmas per sentence, Kuperman                      \\
    208   & at\_AAKuL\_C   & average AoA of lemmas per token, Kuperman                         \\
    209   & to\_AABiL\_C   & total AoA of lemmas, Bird norm                          \\
    210   & as\_AABiL\_C   & average AoA of lemmas, Bird norm per sent           \\
    211   & at\_AABiL\_C   & average AoA of lemmas, Bird norm per token              \\
    212   & to\_AABrL\_C   & total AoA of lemmas, Bristol norm                       \\
    213   & as\_AABrL\_C   & average AoA of lemmas, Bristol norm per sent        \\
    214   & at\_AABrL\_C   & average AoA of lemmas, Bristol norm per token           \\
    215   & to\_AACoL\_C   & total AoA of lemmas, Cortese and Khanna norm                   \\
    216   & as\_AACoL\_C   & average AoA of lem, Cortese and K norm per sent    \\
    217   & at\_AACoL\_C   & average AoA of lem, Cortese and K norm per token       \\ \hline
    \end{tabular}
    }
\caption{\label{Append} Psychollinguistic Features (PsyF)}
\end{table}
\begin{table}[H]
    \centering
    \resizebox{0.5\textwidth}{!}{%
    \begin{tabular}{l@{\hspace{0.3ex}}|l@{\hspace{0.5ex}}|l@{\hspace{0.5ex}}}
    \hline
        \textbf{$idx$} & \textbf{Code}  & \textbf{Definition} \\ \hline
    218   & to\_SbFrQ\_C   & total SubtlexUS FREQcount value                  \\
    219   & as\_SbFrQ\_C   & average SubtlexUS FREQcount value per sentence    \\
    220   & at\_SbFrQ\_C   & average SubtlexUS FREQcount value per token      \\
    221   & to\_SbCDC\_C   & total SubtlexUS CDcount value                    \\
    222   & as\_SbCDC\_C   & average SubtlexUS CDcount value per sent     \\
    223   & at\_SbCDC\_C   & average SubtlexUS CDcount value per token        \\
    224   & to\_SbFrL\_C   & total SubtlexUS FREQlow value                    \\
    225   & as\_SbFrL\_C   & average SubtlexUS FREQlow value per sent     \\
    226   & at\_SbFrL\_C   & average SubtlexUS FREQlow value per token        \\
    227   & to\_SbCDL\_C   & total SubtlexUS CDlow value                      \\
    228   & as\_SbCDL\_C   & average SubtlexUS CDlow value per sent       \\
    229   & at\_SbCDL\_C   & average SubtlexUS CDlow value per token          \\
    230   & to\_SbSBW\_C   & total SubtlexUS SUBTLWF value                    \\
    231   & as\_SbSBW\_C   & average SubtlexUS SUBTLWF value per sent     \\
    232   & at\_SbSBW\_C   & average SubtlexUS SUBTLWF value per token        \\
    233   & to\_SbL1W\_C   & total SubtlexUS Lg10WF value                     \\
    234   & as\_SbL1W\_C   & average SubtlexUS Lg10WF value per sent      \\
    235   & at\_SbL1W\_C   & average SubtlexUS Lg10WF value per token         \\
    236   & to\_SbSBC\_C   & total SubtlexUS SUBTLCD value                    \\
    237   & as\_SbSBC\_C   & average SubtlexUS SUBTLCD value per sent     \\
    238   & at\_SbSBC\_C   & average SubtlexUS SUBTLCD value per token        \\
    239   & to\_SbL1C\_C   & total SubtlexUS Lg10CD value                     \\
    240   & as\_SbL1C\_C   & average SubtlexUS Lg10CD value per sent      \\
    241   & at\_SbL1C\_C   & average SubtlexUS Lg10CD value per token         \\ \hline
    \end{tabular}
    }
\caption{\label{Append} Word Familiarity Features (WorF)}
\end{table}
\begin{table}[H]
    \centering
    \resizebox{0.5\textwidth}{!}{%
    \begin{tabular}{l@{\hspace{0.3ex}}|l@{\hspace{0.3ex}}|l@{\hspace{0.3ex}}}
    \hline
        \textbf{$idx$} & \textbf{Code}  & \textbf{Definition} \\ \hline      
    242   & TokSenM\_S   & total count of tokens x total count of sentence          \\                      
    243   & TokSenS\_S   & sqrt(total count of tokens x total count of sentence)    \\                         
    244   & TokSenL\_S   & log(total count of tokens)/log(total count of sent)  \\                       
    245   & as\_Token\_C  & average count of tokens per sentence                     \\                      
    246   & as\_Sylla\_C  & average count of syllables per sentence                  \\                      
    247   & at\_Sylla\_C  & average count of syllables per token                     \\                      
    248   & as\_Chara\_C  & average count of characters per sentence                 \\                      
    249   & at\_Chara\_C  & average count of characters per token                    \\ \hline                      
    \end{tabular}
    }
\caption{\label{Append} Shallow Features (ShaF)}
\end{table}
\begin{table}[H]
    \centering
    \resizebox{0.5\textwidth}{!}{%
    \begin{tabular}{l@{\hspace{0.3ex}}|l@{\hspace{0.5ex}}|l@{\hspace{0.5ex}}}
    \hline
        \textbf{$idx$} & \textbf{Code}  & \textbf{Definition} \\ \hline      
250   & SmogInd\_S    & Smog Index                        \\
251   & ColeLia\_S    & Coleman Liau Readability Score   \\
252   & Gunning\_S    & Gunning Fog Count Score (New, US Navy Report)           \\
253   & AutoRea\_S    & Automated Readability Idx (New, US Navy Report)   \\ 
254   & FleschG\_S    & Flesch Kincaid Grade Level (New, US Navy Report)        \\ 
255   & LinseaW\_S    & Linsear Write Formula Score       \\ \hline
                    
    \end{tabular}
    }
\caption{\label{Append} Shallow Features (ShaF)}
\end{table}

\section{Rules Behind Feature Codes}
In table 8$\sim$22, ``Code'' columns show feature codes. The related linguistic features appear with quite a number of variations across academia, without a naming convention \citep{A09, 11, A10, A11, A13, A15, A19, A20, lee2020lxper}. For consistency, we set ourselves a few naming rules.
\begin{enumerate}[leftmargin=*]
  \setlength\itemsep{0.1ex}
  \item Feature codes consist of 8 letters/numerals, with 1 or 2 underscores depending on feature types.
  \item All features classify into either count-based or score-based, following popular convention.
  \begin{itemize}[leftmargin=*]
      \item Count-based
      \begin{itemize}[leftmargin=*]
      \item define: final calculation uses simple counts (i.e. total, avg per sent, avg per token, ratio)
      \item format: $xx$\_$xxxxx$\_$\text{C}$. First two letters are ``to'' (total), ``as'' (avg per sent), ``at'' (avg per token), ``ra'' (ratio). Five letters in the middle explain what the feature is. Last letter always ``C.'' Two underscores in between.
      \end{itemize}
      \item Score-based
      \begin{itemize}[leftmargin=*]
      \item define: require additional calculation (e.g. log, square), or famous features with pre-defined names (e.g. Flesch-Kincaid, TTR).
      \item format: $xxxxxxx$\_$\text{S}$. Seven letters are all dedicated to explaining what the feature is. Last letter always ``S.'' One underscore.
      \end{itemize}
  \end{itemize}
  \item For the ``explanation'' part of each feature code, capital letters denote new words. The small letters that follow are from the same word. (e.g. 1: Coleman Liau $\rightarrow$ ColeLia, 2: AoA (Age of Acquisition) Kuperman of words $\rightarrow$ AAKuW)
\end{enumerate}

\section{Scientific Artifacts}
We use Online LDA implemented by Gensim v4.0 \citep{rehurek_lrec}. For most general tasks, including sentence/entity recognition, POS tagging, and dependency parsing, we use spaCy v3.0\footnote{github.com/explosion/spaCy} \citep{spacy} with en\_core\_web\_sm pretrained model. For constituency parsing, we use CRF parser \citep{zhang-etal-2020-fast} in SuPar v1.0 \footnote{github.com/yzhangcs/parser}. 

\subsection{Transformers}
\noindent
For transformers, we use the following models from HuggingFace transformers v4.5.0 \citep{wolf-etal-2020-transformers}.

\noindent
1. \textbf{bert-base-uncased}

\noindent
2. \textbf{roberta-base}

\noindent
3. \textbf{bart-base}

\noindent
4. \textbf{xlnet-base-cased} 

\subsection{Non-Neural Models}
\noindent
For non-neural models, we use the following models from from SciKit-Learn v0.24.1.

\noindent
1. \textbf{support vector classifiers} (svm.SVC) \citep{10.1109/5254.708428, Platt99probabilisticoutputs,chang2011libsvm}

\noindent
2. \textbf{random forest classifiers} (ensemble.RandomF\\orestClassifier) \citep{breiman2001random}

\noindent
3. \textbf{logistic regression} (linear\_model.LogisticRegr\\ession) 

For gradient boosting, we use the following from XGBoost v1.4.0 \citep{chen2016xgboost}.

\noindent
4. \textbf{gradient boosting} (XGBclassifier)

\section{Preprocessing}
Our preprocessing steps are inspired by \citet{Martinc:21} and several other existing RA research. These steps are used only during the extraction of handcrafted features for increased accuracy. 

\noindent
1. remove all special characters

\noindent
2. remove words less than 3 characters

\noindent
3. lowercase all

\noindent
4. tokenize

\noindent
5. remove NLTK default stopwords

\section{Full Hyperparameters}
\begin{table}
\begin{subtable}{0.23\textwidth}
\centering
\resizebox{\textwidth}{!}{%
\begin{tabular}{c|ccc}
\hline
\textbf{Model}& \multicolumn{3}{c}{\textbf{Hyperparameter}} \\ 
\cmidrule(lr){1-1}\cmidrule(lr){2-4}
\multirow{5}{*}{SVM}  & C          & G              & K       \\ 
\cmidrule(lr){2-2}\cmidrule(lr){3-3}\cmidrule(lr){4-4}
                      & \textbf{1} & \textbf{scale} & rbf     \\ 
                      & 5          & auto           & linear  \\
                      & 10         &                & \textbf{poly}    \\
                      & 50         &                & sigmoid \\ \cmidrule(lr){1-4}
\end{tabular}
}
\caption{\textit{SVM}, Best Params }\label{tabla4.1}
\end{subtable}
    \hfil
\begin{subtable}{0.23\textwidth}
\centering
\resizebox{\textwidth}{!}{%
\begin{tabular}{c|ccc}
\hline
\textbf{Model}& \multicolumn{3}{c}{\textbf{Hyperparameter}} \\ 
\cmidrule(lr){1-1}\cmidrule(lr){2-4}
\multirow{5}{*}{RF}  & nEst           & MDep          & Mfea       \\ 
\cmidrule(lr){2-2}\cmidrule(lr){3-3}\cmidrule(lr){4-4}
                      & 600           & 20            & \textbf{auto}     \\ 
                      & 700           & 60            & sqrt  \\
                      & \textbf{800}  & 100           & log2    \\
                      & 900           & \textbf{None} & None \\ \cmidrule(lr){1-4}
\end{tabular}
}
\caption{\textit{RandomF}, Best Params }\label{tabla4.1}
\end{subtable}
\newline
\vspace*{1ex}
\newline
\begin{subtable}{0.23\textwidth}
\centering
\resizebox{\textwidth}{!}{%
\begin{tabular}{c|ccc}
\hline
\textbf{Model}& \multicolumn{3}{c}{\textbf{Hyperparameter}} \\ 
\cmidrule(lr){1-1}\cmidrule(lr){2-4}
\multirow{5}{*}{XGBoost}  & eta       & G          & MDep      \\ 
\cmidrule(lr){2-2}\cmidrule(lr){3-3}\cmidrule(lr){4-4}
                      & 1e-2          & 0          & 3      \\ 
                      & \textbf{5e-2}   & 1e-2     & 6   \\
                      & 1e-1          & 1e-1       & \textbf{9}  \\
                      & 2e-1          & \textbf{1} & 12 \\ \cmidrule(lr){1-4}
\end{tabular}
}
\caption{\textit{XGBoost}, Best Params }\label{tabla4.1}
\end{subtable}
    \hfil
\begin{subtable}{0.23\textwidth}
\centering
\resizebox{\textwidth}{!}{%
\begin{tabular}{c|ccc}
\hline
\textbf{Model}& \multicolumn{3}{c}{\textbf{Hyperparameter}} \\ 
\cmidrule(lr){1-1}\cmidrule(lr){2-4}
\multirow{5}{*}{LR}  & C           & Pen          & Solver       \\ 
\cmidrule(lr){2-2}\cmidrule(lr){3-3}\cmidrule(lr){4-4}
                      & 1e-1       & \textbf{l1}  & lbfgs     \\ 
                      & 5e-1       & l2           & l.linear  \\
                      & \textbf{1} & elastic      & newton    \\
                      & 10         & none         & \textbf{saga} \\ \cmidrule(lr){1-4}
\end{tabular}
}
\caption{\textit{LogR}, Best Params }\label{tabla4.1}
\end{subtable}
\caption{Hyperparameters, non-neural models.}
\end{table}
\subsection{Machine Learning Models}
We perform grid search on the hyperparameters (table 3) after performing a large randomized search to identify the sensible range of hyperparameters to tune. In particular, logistic regression solver hyperparameter search include libfgs \citep{zhu2011bfgs}, liblinear \citep{fan2008liblinear}, SAG \citep{schmidt2017minimizing}, and SAGA \citep{defazio2014saga}. 

In table 3(a), C is the regularization parameter, G is the kernel coefficient (gamma), and K is the kernel. In table 3(b), nEst is the number of trees, MDep is the max depth of a tree, and Mfea is the number of features considered. In table 3(c), eta is the learning rate, G is the minimum loss reduction need to make a further partition on the leaf node (gamma), and MDep is the max depth of a tree. In table 4(d), C is the inverse of the regularization strength, Pen is the norm used in penalization, and Solver is the algorithm used in optimization. The other parameters best performed at default.

\subsection{Transformers}
We use AdamW (optimizer) \citep{kingma2014adam}, linear (scheduler), 10\% (warmup steps), 8 (batch size), 3 (epoch) for all tested transformers. We use the learning rate of 2e-5 for BERT and 3e-5 for the other three transformers.

\section{Feature Combinations}
\begin{table}[H]
\resizebox{0.5\textwidth}{!}{%
\begin{tabular}{@{\hspace{0.7ex}}c@{\hspace{0.7ex}}|@{\hspace{0.7ex}}l@{\hspace{0.7ex}}|@{\hspace{0.7ex}}c@{\hspace{0.7ex}}c@{\hspace{0.7ex}}}
\hline
\textbf{Set} & \textbf{Features}                      & \textbf{LogR}&\textbf{SVM} \\ 
\cmidrule(lr){1-1}\cmidrule(lr){2-2}\cmidrule(lr){3-3}\cmidrule(lr){4-4}
\textbf{T1}  & \textbf{AdSem + Disco + Synta + LxSem + ShaTr}  & \textbf{0.622}      & \textbf{0.679}                    \\
T2           & Disco + Synta + LxSem + ShaTr          & 0.528      & 0.546                      \\
T3           & AdSem + Synta + LxSem + ShaTr          & 0.591      & 0.582                      \\
\cmidrule(lr){1-4}
\textbf{H1}           & \textbf{AdSem + Disco}                          & \textbf{0.463}      & \textbf{0.513}                       \\
\cmidrule(lr){1-4}
L1           & Synta + LxSem                          & 0.499      & 0.561                      \\
\textbf{L2}  & \textbf{Set L1 - PhrF}                       & \textbf{0.539}      & \textbf{0.577}                      \\
L3           & Set L1 - VarF                       & 0.529      & 0.551                      \\
L4           & Set L1 - POSF                       & 0.449      & 0.551                      \\
\cmidrule(lr){1-4}
E1           & AdSem + PsyF + WorF + TraF             & 0.489      & 0.473                    \\
\textbf{E2}  & \textbf{AdSem} + \textbf{PsyF} + \textbf{WorF}                    & \textbf{0.490}      & \textbf{0.479}                    \\
E3           & PsyF + WorF                            & 0.464      & 0.459                    \\
\cmidrule(lr){1-4}
P1           & EnDF + ShaF + TrSF + POSF + WorF + PsyF + TraF & 0.608      & 0.633                    \\
P2           & Set P1 + TraF                                 & 0.629      & 0.638                    \\
\textbf{P3}  &\textbf{Set P2 + VarF}                                 & \textbf{0.647}      & \textbf{0.674}                    \\
  \cmidrule(lr){1-4}
\end{tabular}
}
\begin{tablenotes}[para,flushleft]
\small
\item[*]\textit{Note: 5 letters (e.g. AdSem) mean linguistic branch. 4 letters (e.g. PhrF) mean subgroup. We report accuracy on WeeBit.}
\end{tablenotes}
\caption{\label{Table 1} Defining feature sets.}
\end{table}
The five types of feature sets have varying aims: 1. \textbf{T-type} thoroughly captures linguistic properties, 2. \textbf{H-type} captures the high-level properties, 3. \textbf{L-type} captures the low, surface-level properties, 4. \textbf{E-type} uses features calculated from external data (out-of-model info, i.e. Age-of-Acquisition), and 5. \textbf{P-type} collects features by performance. Both advanced semantic and discourse features add distinctive information. This can be evidenced by the performance decreases (T1 $\rightarrow$ T2 and T1 $\rightarrow$ T3). We checked that all measures of F1, precision, recall, and QWK followed the same trend. Similar method was used in \citet{Feng:09}; \citet{aluisio2010readability}; \citet{Vajjala:12}; \citet{falkenjack2013features}; \citet{franccois2014analysis} to check if a feature added orthogonal information. More linguistic branches generally indicated better performance. 
We use SciKit-learn \citep{pedregosa2011scikit} for metrics.
\section{Transformers Training Time}
All numbers are in seconds. We report in the order of (BERT, RoBERTa, XLNet, BART). These are the average training times for each fold, with 80\% of the full dataset used to train. We used an NVIDIA Tesla V100 GPU.

\noindent
1. \textbf{WeeBit} (1546, 1485, 3617, 1202) 

\noindent
2. \textbf{OneStopEnglish} (451, 373, 977, 396)

\noindent
3. \textbf{Cambridge} (215, 122, 393, 239)

\section{LingFeat}
Throughout our paper, we mention LingFeat as one of our contributions to academia. This is because a large-scale handcrafted features extraction toolkit is scarce in RA, despite its reliance on the features. 

LingFeat is a Python research package for various handcrafted linguistic features. More specifically, LingFeat is an NLP feature extraction software, which currently extracts 255 linguistic features from English string input. The package is available on both PyPI and GitHub. 

Due to the wide number of supported features, we had to define subgroups (section 3) for features. Hence, features are not accessible individually. Instead, one has to call the subgroups to obtain the dictionary of the corresponding features. The corresponding code is applicable to LingFeat v.1.0.

\newpage
\begin{lstlisting}[language=Python, basicstyle=\fontsize{8}{10}\selectfont\ttfamily]
"""
Import

this is the only import you need
"""
from lingfeat import extractor


"""
Pass text

here, text must be in string type
"""
text = "..."
LingFeat = extractor.pass_text(text)


"""
Preprocess text

options (all boolean):
- short (def. False): include short words
- see_token (def. False): return token list
- see_sent_token (def. False): return sent

output:
- n_token
- n_sent
- token_list (optional)
- sent_token_list (optional)
"""
LingFeat.preprocess()
# or
# print(LingFeat.preprocess())


"""
Extract features

each method returns a dictionary of 
the corresponding features
"""
# Advanced Semantic (AdSem) Features
WoKF=LingFeat.WoKF_() #Wiki Know. Features
WBKF=LingFeat.WBKF_() #WB Knowledge Features
OSKF=LingFeat.OSKF_() #OSE Knowledge Features

# Discourse (Disco) Features
EnDF=LingFeat.EnDF_() #Entity Dens. Features
EnGF=LingFeat.EnGF_() #Entity Grid Features

# Syntactic (Synta) Features
PhrF=LingFeat.PhrF_() #Phrasal Features
TrSF=LingFeat.TrSF_() #(Parse) Tree Features
POSF=LingFeat.POSF_() #POS Features

# Lexico Semantic (LxSem) Features
TTRF=LingFeat.TTRF_() #TTR Features
VarF=LingFeat.VarF_() #Variational Features
PsyF=LingFeat.PsyF_() #Psycholing Difficulty 
WoLF=LingFeat.WorF_() #Word Familiarity

# Shallow Traditional (ShTra) Features
ShaF=LingFeat.ShaF_() #Shallow Features 
TraF=LingFeat.TraF_() #Traditional Formulas 
\end{lstlisting}
\end{document}